\documentclass{article}

\usepackage{microtype}
\usepackage{graphicx}
\usepackage{subfigure}
\usepackage{booktabs} 
\usepackage{url}
\usepackage{float}
\usepackage{floatflt}
\usepackage{amsmath}
\usepackage{bm}
\usepackage{tikz}
\usepackage{ctable}
\usepackage{multicol}
\usepackage{multirow}
\usepackage{subcaption}
\usepackage{wrapfig}
\usepackage{xcolor}
\usepackage{wrapfig}
\usepackage{algorithm}
\usepackage{colortbl}
\usepackage{pifont}

\usepackage{tabularx}
\usepackage[makeroom]{cancel}

\usepackage{microtype}
\usepackage{graphicx}
\usepackage{subfigure}
\usepackage{physics}
\usepackage{booktabs} 
\usepackage{colortbl}
\def\redc{\cellcolor[HTML]{FF999A}}
\def\orangec{\cellcolor[HTML]{FFCC99}}
\def\yellowc{\cellcolor[HTML]{FFF8AD}}

\usepackage{hyperref}
\def\our{RaySplats}
\usepackage{amsmath}

\def\R{\mathbb{R}}

\def\L{\mathcal{L}}
\def\G{\mathcal{G}}
\def\N{\mathcal{N}}
\def\m{\mathrm{m}}

\def\E{\mathcal{E}}

\usepackage{hyperref}

 \usepackage[accepted]{icml2024}

\usepackage{amsmath}
\usepackage{amssymb}
\usepackage{mathtools}
\usepackage{amsthm}
\usepackage{soul}
\usepackage[normalem]{ulem}
\usepackage[T1]{fontenc}

\usepackage[capitalize,noabbrev]{cleveref}

\theoremstyle{plain}
\newtheorem{theorem}{Theorem}[section]
\newtheorem{proposition}[theorem]{Proposition}

\theoremstyle{definition}

\theoremstyle{remark}

\usepackage[textsize=tiny]{todonotes}

\def\R{\mathbb{R}}

\begin{document}

\twocolumn[
\icmltitle{ 
\our{}: Ray Tracing based Gaussian Splatting
}



\icmlsetsymbol{equal}{*}

\begin{icmlauthorlist}
\icmlauthor{Krzysztof Byrski}{yyy}
\icmlauthor{Marcin Mazur}{yyy}
\icmlauthor{Jacek Tabor}{yyy}
\icmlauthor{Tadeusz Dziarmaga}{yyy}
\icmlauthor{Marcin K\k{a}dziołka}{yyy}
\icmlauthor{Dawid Baran}{yyy}
\icmlauthor{Przemys\l{}aw Spurek}{yyy}

\end{icmlauthorlist}
\icmlaffiliation{yyy}{Jagiellonian University, Faculty of Mathematics
and Computer Science, Cracow, Poland}

\icmlcorrespondingauthor{Przemys\l{}aw Spurek}{przemyslaw.spurek@uj.edu.pl}

\icmlkeywords{Gaussian Splatting, Mesh, NeRF, Rendering}

\vskip 0.3in
]




\printAffiliationsAndNotice{\icmlEqualContribution} 

\begin{abstract}
3D Gaussian Splatting (3DGS) is a process that enables the direct creation of 3D objects from 2D images. This representation offers numerous advantages, including rapid training and rendering. However, a significant limitation of 3DGS is the challenge of incorporating light and shadow reflections, primarily due to the utilization of rasterization rather than ray tracing for rendering. This paper introduces \our{}, a model that employs ray-tracing based Gaussian Splatting. Rather than utilizing the projection of Gaussians, our method employs a ray-tracing mechanism, operating directly on Gaussian primitives represented by confidence ellipses with RGB colors. In practice, we compute the intersection between ellipses and rays to construct ray-tracing algorithms, facilitating the incorporation of meshes with Gaussian Splatting models and the addition of lights, shadows, and other related effects.  \url{https://github.com/KByrski/RaySplatting}
\end{abstract}

\section{Introduction}

3D Gaussian Splatting (3DGS) \cite{kerbl20233d} is a leading neural rendering technique capable of transforming 2D images into coherent 3D scenes with remarkably sharp reconstruction. 3DGS uses a collection of Gaussians characterized by color and opacity to represent scenes, increasing training efficiency and enabling real-time rendering of high-quality meshes. The rendering process involves projecting these 3D components onto a 2D plane. While numerically efficient, it has certain limitations, particularly in terms of integrating 3DGS with objects based on meshes and lighting effects, a process that poses significant challenges, primarily due to the complexities associated with ray tracing \cite{glassner1989introduction}. 

\begin{figure}[!t]
    \centering
        \begin{center}
            \our{} for Glass Objects with Shadows\\
        \end{center}
        \includegraphics[width=\linewidth]{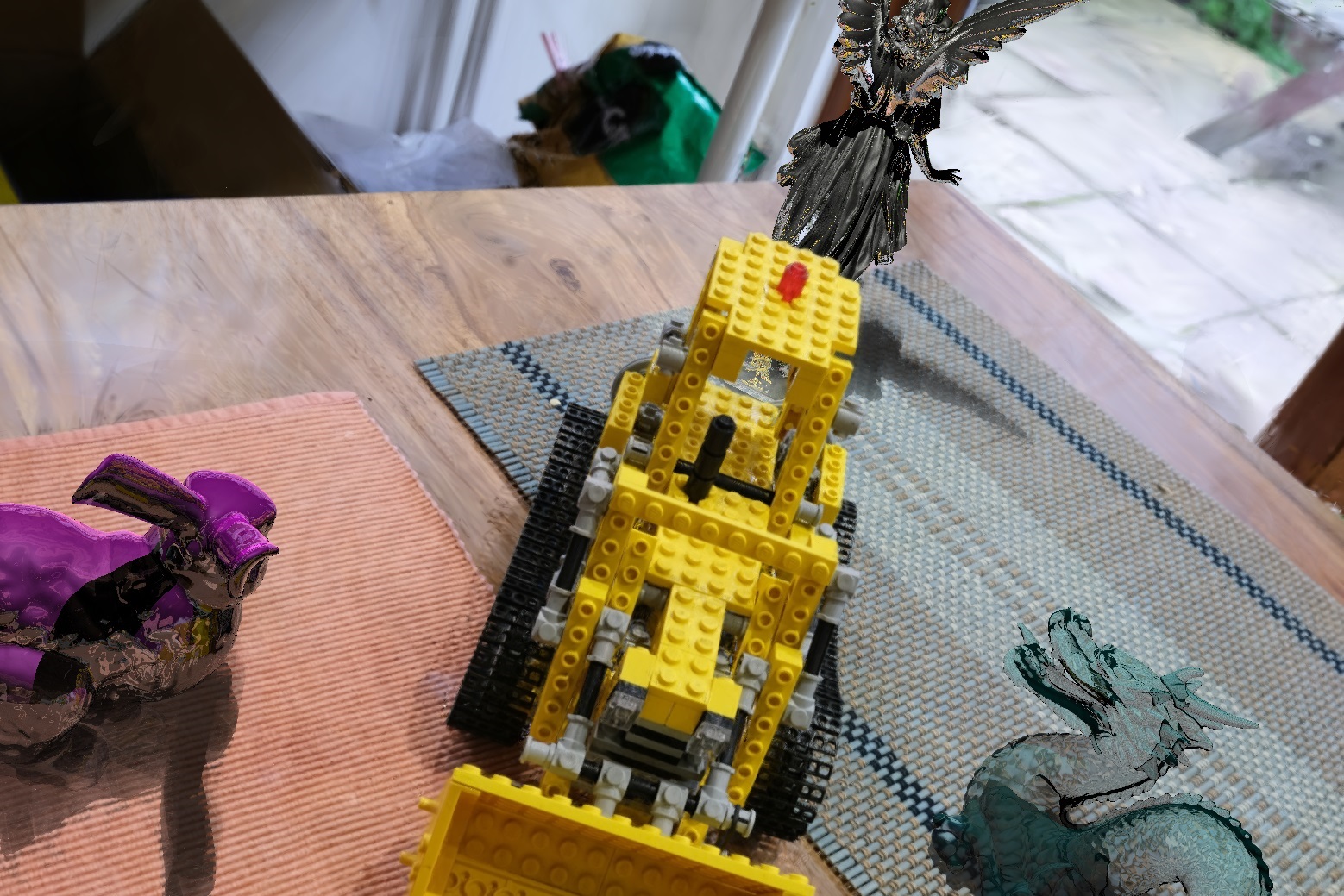}\\
            \our{} for Mirror Reflections\\[1mm]
        \includegraphics[width=\linewidth]{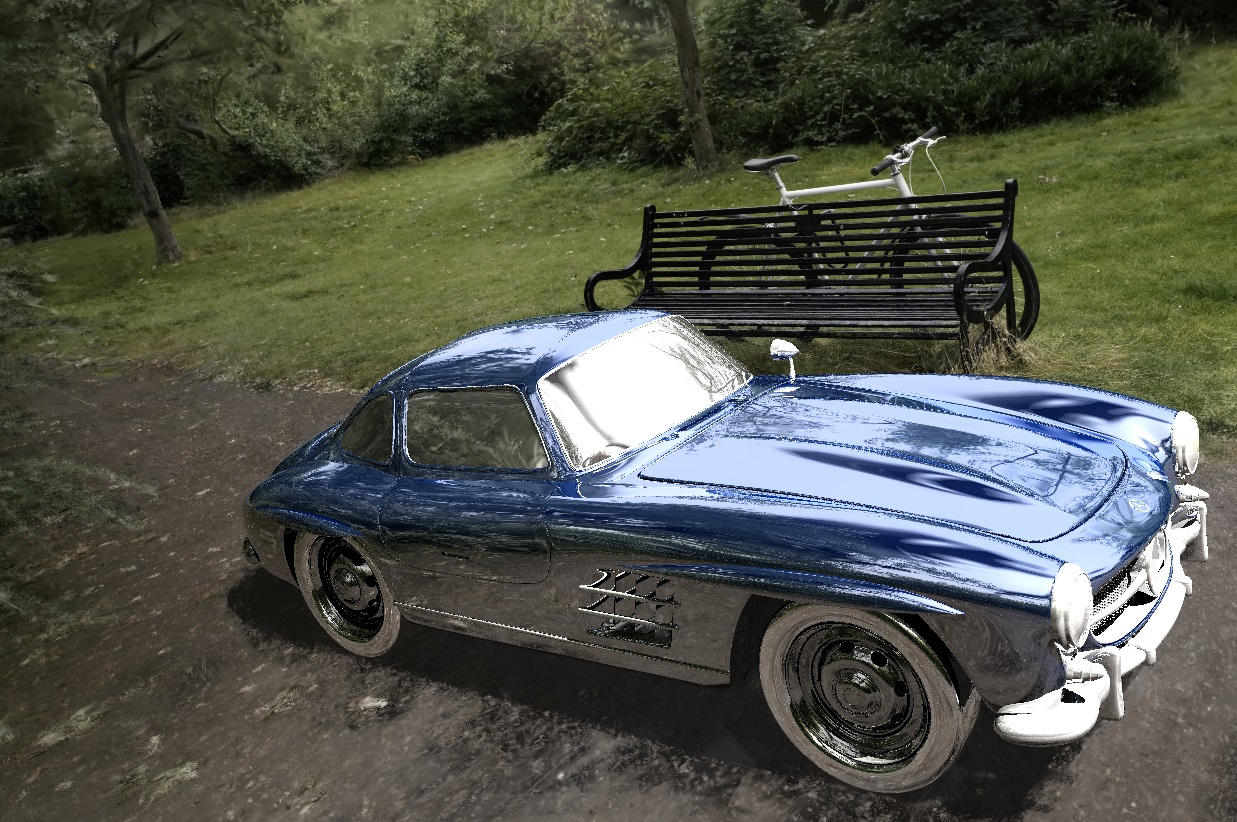}
    \caption{\our{} (our) incorporates ray tracing into the 3D~Gaussian Splatting framework. This allows us to integrate meshes with lighting conditions and mirror effects.}
    \label{fig:trajectorid}
\end{figure}

3D Gaussian Ray Tracing (3DGRT) integrates the principles of 3D Gaussian Splatting with the ray-tracing model, leveraging bounding primitives (polytopes) for each Gaussian. This approach utilizes iterations and is efficient, producing high-quality renders thanks to the NVIDIA OptiX programming interface \cite{parker2010optix}. However, 3DGRT involves the approximation of primitives. This technique can be challenging, especially for flat Gaussians, which are often used to generate meshes from 3DGS. Inter-Reflective Gaussian Splatting (IRGS) \cite{gu2024irgs} integrates flat Gaussians from 2DGS \cite{huang20242d} to effectively capture inter-reflections. Utilizing the rendering equation, IRGS proposes a differentiable 2D Gaussian ray-tracing approach within the 2DGS framework, thereby facilitating re-lighting modeling within 2DGS.

As demonstrated above, although both 3DGRT and 2DGS are capable of producing high-quality renders, they are subject to certain limitations. Specifically, 3DGRT utilizes bounding primitives (polytopes), while IRGS is exclusively dedicated to relighting 2DGS.
In order to address this issue, we introduce Ray Tracing based Gaussian Splatting (\our{})\footnote{The source code is available at \url{https://github.com/KByrski/RaySplatting}.}, a model that utilizes ray tracing during both training and inference. \our{} eschews the use of the polytope approximation and is compatible with 2D cases involving flat Gaussians. In practice, we calculate the distance between Gaussians represented by ellipses with rays. This approach facilitates the integration of our model with light conditions and the merging of scenes with mesh-based objects.

The following is a concise summary of our contribution:
\begin{itemize}
    \item  we propose \our{}, a novel differential rendering procedure for 3D Gaussian Splatting, which utilizes ray tracing during both the training and inference phases, yet does not necessitate the use of polytope approximation,
    \item \our{} enables the processing of lighting effects, including reflections, shadows, and transparency,
    \item \our{} facilitates the integration of 3D Gaussian Splatting within mesh-based models, thereby enabling the application of 2D Gaussians.
\end{itemize}

\section{Related Works}

This section is divided into two parts. First, we describe novel-view synthesis in general. Then, we discuss ray-tracing methods.

\paragraph{Novel-View Synthesis}
Neural Radiance Fields (NeRFs) \cite{mildenhall2021nerf} have significantly advanced the field of novel-view synthesis by representing scenes as a volumetric radiance field encoded in a coordinate-based neural network. NeRFs allow the network to be queried at any point to retrieve volumetric density and view-dependent color, resulting in photorealistic scene renderings. Due to their high-quality output, NeRFs have become a foundational approach for novel view synthesis.

\begin{figure}
    \centering
        \includegraphics[width=0.9\linewidth]{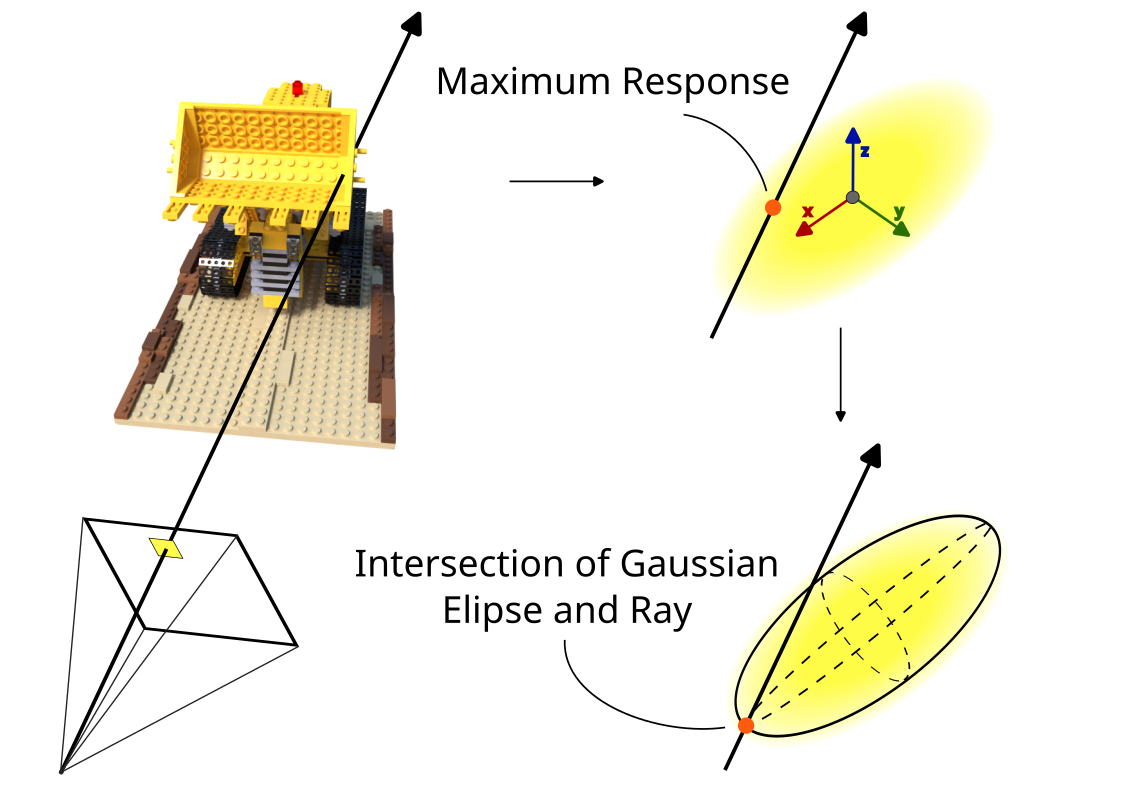}
    \caption{\our{} (our) uses ray-tracing based solutions. In practice, we need two important points on rays passing through Gaussian distributions. Then, the maximum response point is utilized for aggregating colors along each ray. On the other hand, the intersection of Gaussian confidence ellipses is used to efficiently detect Gaussians with non-empty intersection with the ray.}
    \label{fig:ray_elipses}
\end{figure}

Building on NeRF’s success, 3D Gaussian Splatting (3DGS) \cite{kerbl20233d} introduced a novel representation of scenes using fuzzy, anisotropic 3D Gaussian point clouds. 
This technique utilizes a tile-based rasterizer to project 3D Gaussians onto a 2D plane, achieving state-of-the-art results in terms of visual fidelity and rendering efficiency. The efficacy of 3DGS has been demonstrated in a variety of applications, including geometry reconstruction \cite{kerbl20233d}, dynamic scene reconstruction \cite{yang2023real,waczynska2024d}, inverse rendering \cite{jiang2024gaussianshader,liang2024gs}, and street scene visualization \cite{chen2023periodic}. However, relying on rasterization makes it harder for Gaussian Splatting to accurately replicate ray-based effects \cite{moenne20243d}.

\begin{figure*}
    \centering
        \begin{center}
            \our{} for Glass Objects with Shadows\\
        \end{center}
        \includegraphics[width=0.32\linewidth]{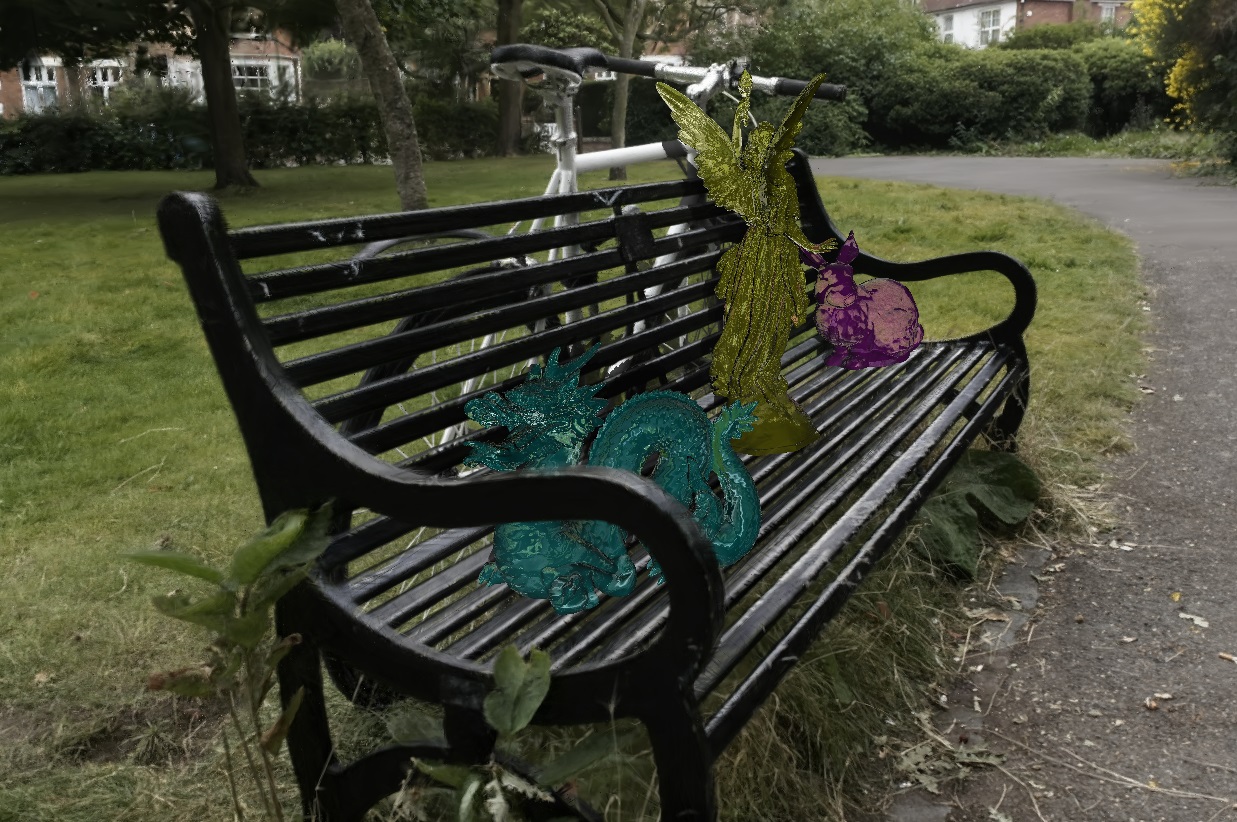}
        \includegraphics[width=0.32\linewidth]{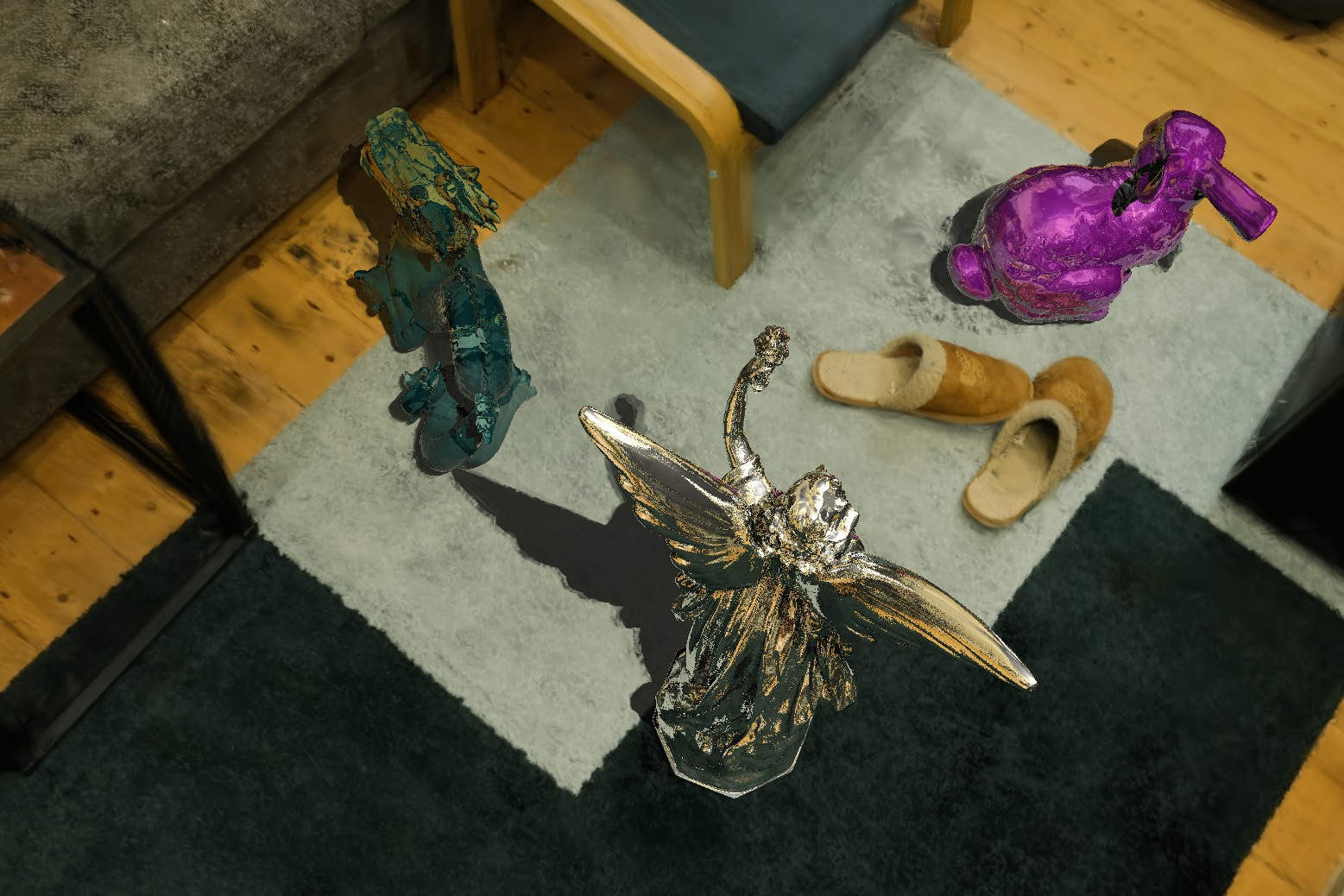}
        \includegraphics[width=0.329\linewidth]{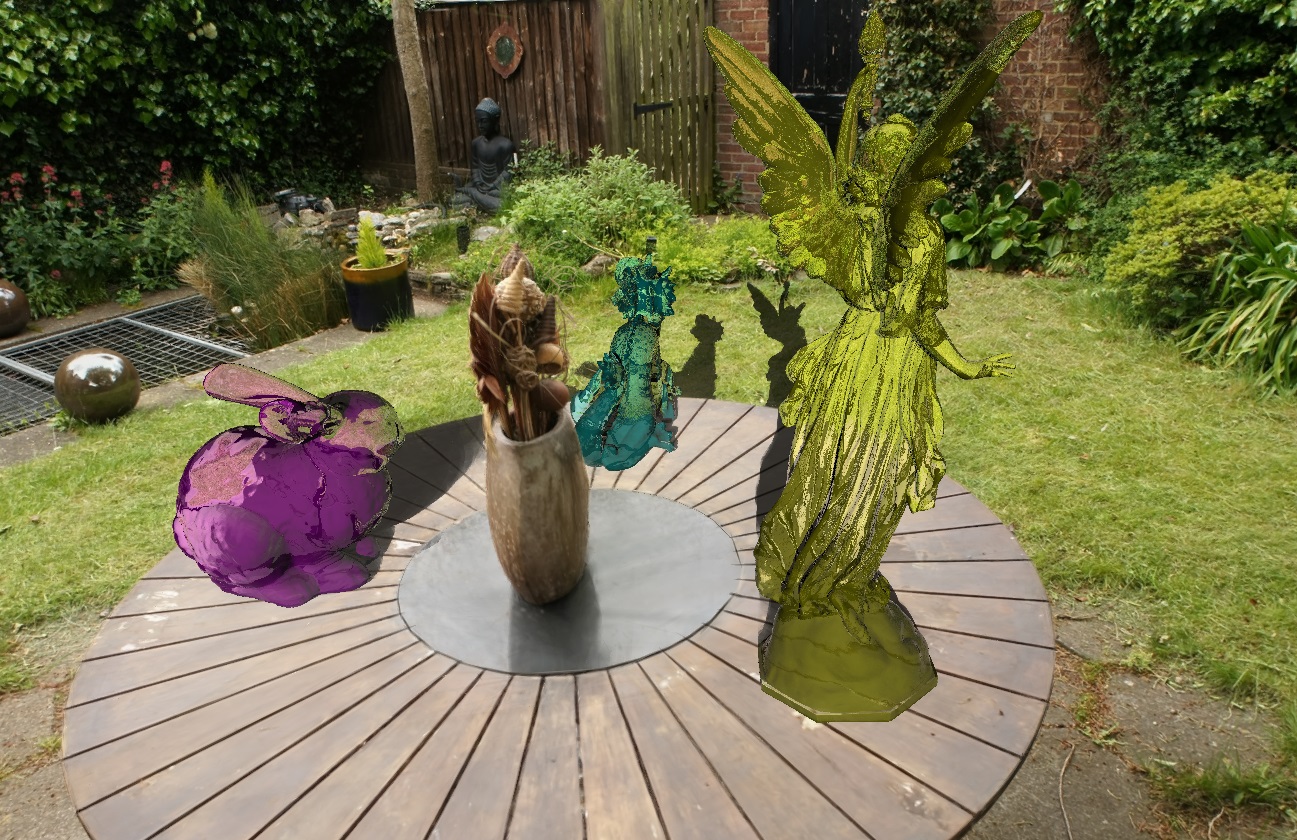}
    \caption{\our{} (our) allows us to combine 3D Gaussian splatting with a mesh-based rendering using lighting effects such as shadows and transparency.}
    \label{fig:3glasses}
\end{figure*}

\begin{figure*}
    \centering
        \begin{center}
            \our{} for Mirror Reflections\\
        \end{center}
        \includegraphics[width=0.33\linewidth, trim=0 100 0 0 , clip]{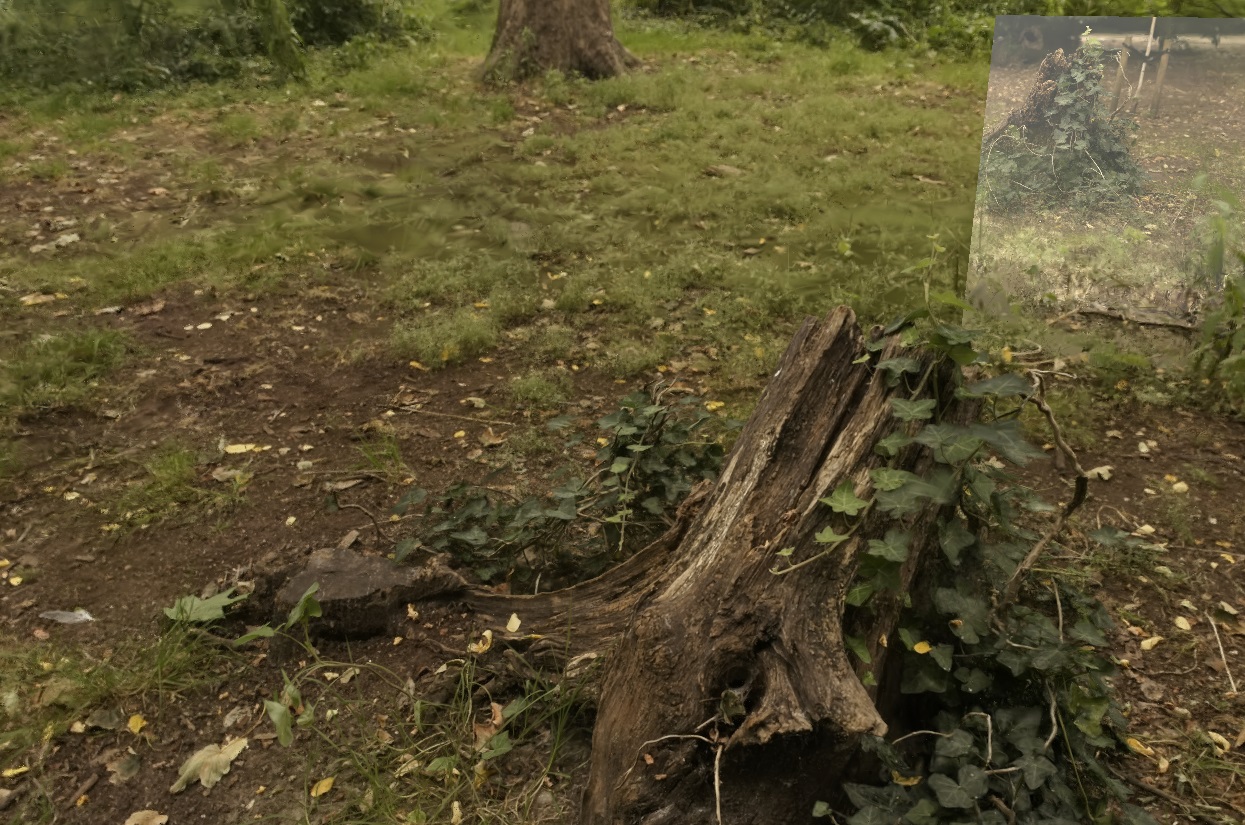}      \includegraphics[width=0.33\linewidth, trim=0 100 0 0 , clip]{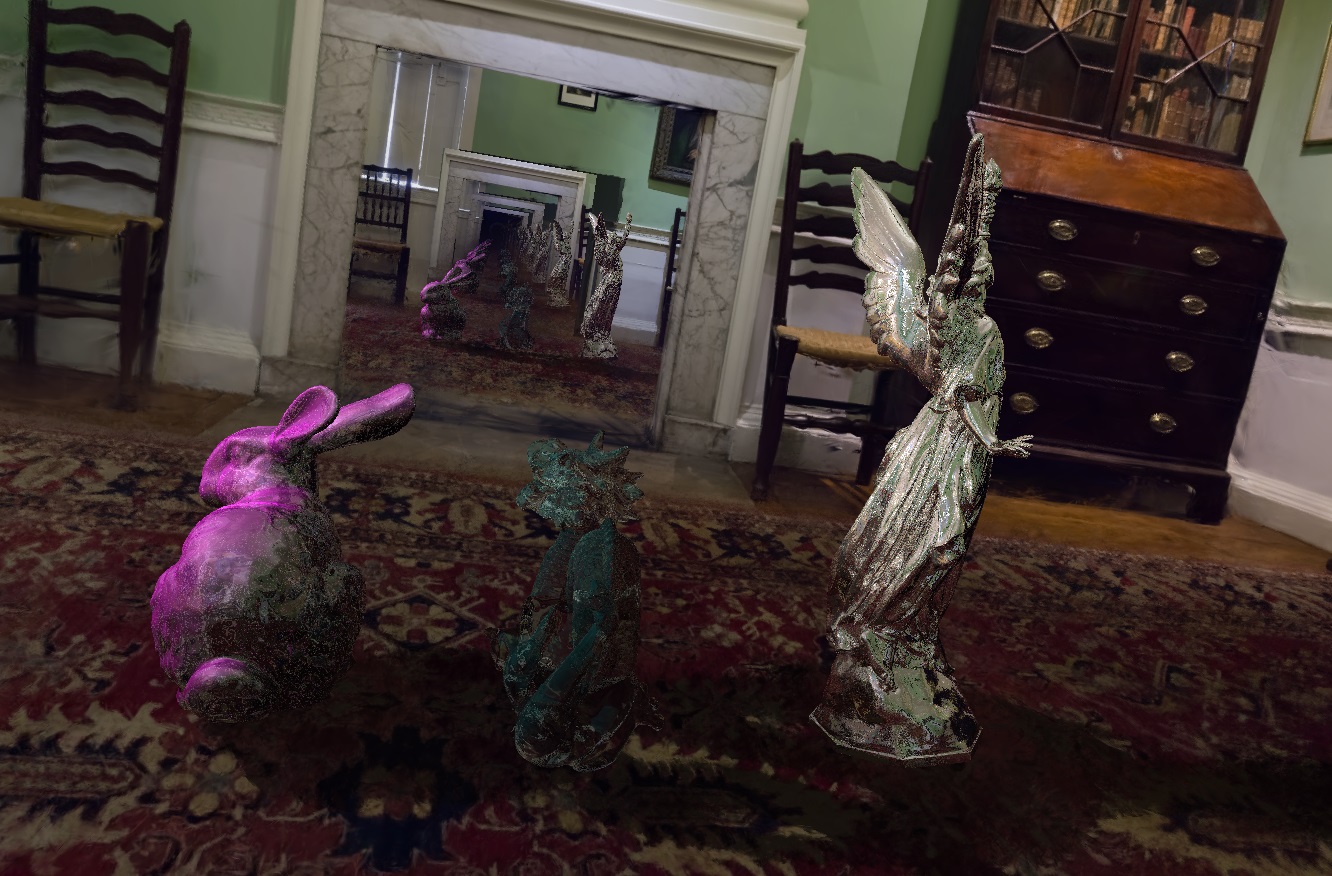}
        \includegraphics[width=0.33\linewidth]{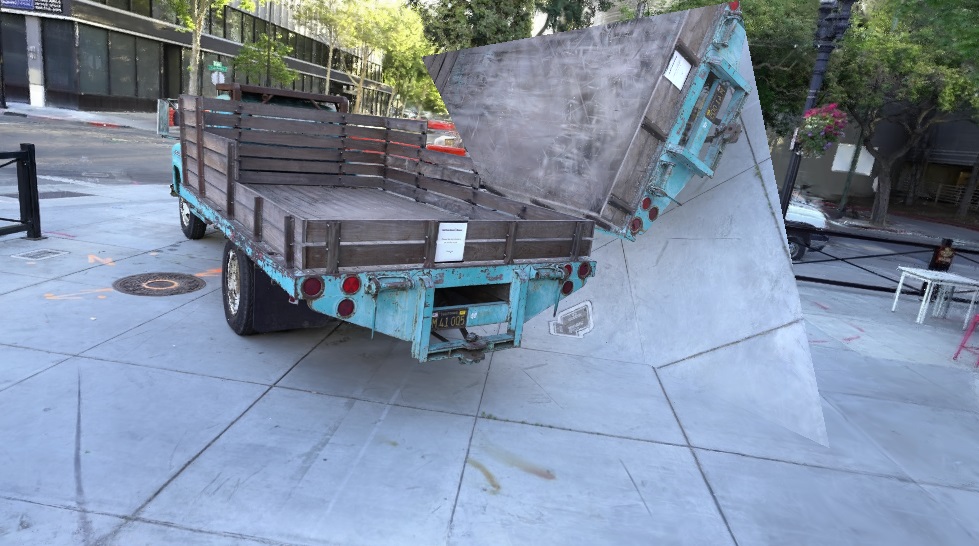}
    \caption{\our{} (our) allows us to combine 3D Gaussian splatting with a mesh-based rendering using lighting effects such as mirror reflections.}
    \label{fig:mirror}
\end{figure*}

\paragraph{Ray-Tracing Methods}
To address the aforementioned limitation, 3D Gaussian Ray Tracing (3DGRT) \cite{moenne20243d} introduced a differentiable ray tracer for 3D Gaussian primitives. This method efficiently performs ray tracing across 3D
Gaussian primitives, allowing radiance computation along arbitrary rays. Utilizing the NVIDIA OptiX programming interface \cite{parker2010optix}, 3DGRT achieves high-quality renders with improved accuracy over rasterization-based approaches. However, its scope remains constrained to Gaussian primitives, limiting its compatibility with generalized distributions \cite{condor2025don,hamdi2024ges,kasymov2024neggs}.

Subsequent advancements have focused on particular challenges inherent to Gaussian-based rendering. For example, Inter-Reflective Gaussian Splatting (IRGS) \cite{gu2024irgs} extends the 2D Gaussian Splatting (2DGS) framework \cite{huang20242d} to model inter-reflections by incorporating a differentiable ray-tracing approach. By integrating flat Gaussians from 2DGS with the rendering equation, IRGS facilitates relighting modeling. While IRGS achieves high-quality results, its emphasis on 2D Gaussian primitives restricts its applicability to 3D scene reconstruction or integration with mesh-based models.
Another approach, EnvGS \cite{xie2024envgs}, introduces a ray tracing-based approach to model complex reflections in real world scenes. By employing environment Gaussian primitives for near-field and high-frequency reflections, it overcomes the limitations of environment maps, which struggle with accurate reflection modeling due to distant lighting assumptions. EnvGS integrates these environment Gaussians with base Gaussian primitives, enabling detailed, real-time rendering. Unfortunately, this method involves a two-step optimization process, where, in practice, ray tracing is applied solely to the previously trained 3DGS.

Despite advancements, challenges persist in integrating Gaussian representations with meshes and simulating lighting effects like reflections, shadows, and transparency. Our \our{} method overcomes these difficulties by extending the capabilities of 3DGS to include lighting effects and allowing seamless integration with mesh-based models.




\begin{figure}[t]
    \centering
        \begin{center}
            \our{} for Glass Objects with Shadows\\[-5mm]
        \end{center}
        \includegraphics[width=0.95\linewidth, trim=0 80 0 50 , clip]{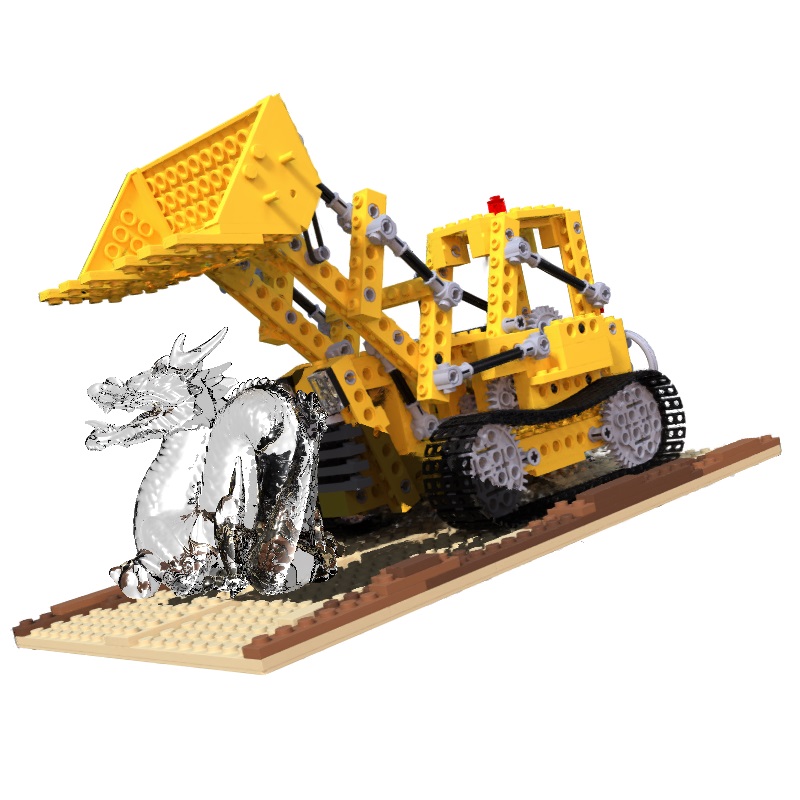}        
    \caption{\our{} (our) is capable of modeling glass elements in the 3D Gaussian Splatting environment, thereby facilitating the accurate visualization of glass reflections and the distortion of light due to refraction.}
    \label{fig:glass}
\end{figure}

\section{\our{} -- Ray Tracing based Gaussian Splatting }

This section is intended to provide an overview of our model. First, the classical 3D Gaussian Splatting approach is discussed. Next, the process of determining the intersection of a ray with a Gaussian distribution is outlined. It is important to note that identifying Gaussians that intersect a given ray is crucial for aggregating colors along that ray. This concept is subsequently elaborated upon in great detail. Finally, the loss function of the \our{} model is presented.

\paragraph{3D Gaussian Splatting} 
3D Gaussian Splatting (3DGS) \cite{kerbl20233d} is a method of representing 3D scenes as a set of Gaussians
\begin{equation}\label{eq:3dgs}
    \G = \{ (\N(\m_i,\Sigma_i), \hat{\alpha}_i, c_i) \}_{i=1}^{n},
\end{equation}
where trainable constants $m_i$, $\Sigma_i$, $\hat{\alpha}_i$, and $c_i$ denote the position (mean), covariance, opacity, and color of the $i$-th component. The representation of color employs the Spherical Harmonics (SH) approach \cite{fridovich2022plenoxels,muller2022instant}. 3DGS utilizes the factorization of the covariance matrix
\begin{equation}\label{eq:factorization}
    \Sigma = RSSR^T,
\end{equation}
where $R$ is the rotation matrix and $S$ is a diagonal matrix containing the scaling parameters. 

The 3DGS algorithm entails the projection of Gaussians onto the image plane, wherein the color for a specific pixel is determined by the blending of a sequence of overlapping points, which are sampled from the respective Gaussian distributions $\N(m_i, \Sigma_i)$ (we assume that these are indicated by indices $i\in \{1,\ldots,N\}$). This procedure is analogous to that described \cite{kopanas2021point, kopanas2022neural}. The color of a pixel is thus determined as follows:
\begin{equation}\label{equ:color}
C = \sum_{i=1}^N c_i \alpha_i \prod_{j=1}^{i-1} (1-\alpha_j)
\end{equation}
where $\alpha_i$ is given by evaluating a 2D Gaussian with covariance matrix $\Sigma$  multiplied by a learned per-Gaussian opacity $\hat{\alpha}_i$ \cite{yifan2019differentiable}.

The aforementioned formula is a consequence of point-based alpha blending \cite{kerbl20233d}, wherein the color of a ray is defined as
\begin{equation}\label{eq:color1}
    C = \sum_{i=1}^N T_i (1-\exp(-\sigma_i\delta_i))c_i
\end{equation}
with 
\begin{equation}\label{eq:color2}
T_i = \exp(-\sum_{j=1}^{i-1} \sigma_j \delta_j),
\end{equation}
and samples of density $\sigma_i$, transmittance $T_i$, and color $c_i$ are collected along the ray with intervals $\delta_i$. It is imperative to acknowledge the primary challenge in the parametrization of each Gaussian, which encompasses its position, a covariance matrix ($3\times 3$ matrix), color, and opacity. On the other hand, neural rendering is substituted by projecting Gaussian distributions onto a 2D plane. This process is efficient in training and inference; however, issues emerge in the context of light conditioning. To address this challenge, we propose a solution that integrates 3DGS with ray tracing.

\paragraph{Intersection of the Gaussian Distribution with the Ray}
\label{intersection}

We use approaches similar to DSS~\cite{yifan2019differentiable}, which describes a high-fidelity differentiable renderer for point clouds.
The expected color $C({\bf r})$ of the camera ray, defined as ${\bf r}(t) = {\bf o} + t {\bf d}$ (where ${\bf o}$ represents the origin and ${\bf d}$ the direction), is determined by combining the colors and opacities of Gaussians intersected by the ray. Therefore, it is important to define how the Gaussians and the ray intersect.  
The Gaussian component can be thought of as an ellipse. Suppose a random vector ${\bf x}$ follows a 3D Gaussian distribution. The Mahalanobis distance of the vector ${\bf x}$ from the mean vector ${\bf \mu}$ is denoted by the scalar quantity $({\bf x}-{\bf \mu})^T\Sigma^{-1}({\bf x}-{\bf \mu})$. The surface on which the value of the random variable is constant forms an ellipse (or an ellipsoid in a multivariate context) with its center at $\mu$. This ellipse, also known as a probability contour, describes the minimum region (or volume in a multivariate situation) that contains a given probability under the assumption of a Gaussian distribution. For the confidence level $\alpha \in [0,1]$, we can compute the confidence ellipsoid
\begin{equation}
    \E_{{\bf \mu}, \Sigma, \alpha} = \{ {\bf x} \in \R^3 \colon ({\bf x}-{\bf \mu})^T\Sigma^{-1}({\bf x}-{\bf \mu}) = Q\}.
\end{equation}
where $Q = F_{\chi^2(3)}^{-1} \left( \alpha \right)$ is the quantile of order $\alpha$ of the $\chi^2(3)$ distribution (i.e., with three degrees of freedom). Our rendering procedure uses a ``2.5D'' approach using a point from a Gaussian confidence ellipse. In practice, we take the point from the intersection of the ellipses and the ray closest to the camera position. If the line does not hit the Gaussian at all, or if the closest intersection point (i.e., the one with the smaller value of the $t$ parameter) is negative and does not belong to the ray, we treat that ray as the ray missing the Gaussian. In our implementation, instead of the confidence level $\alpha$, for flexibility we use the configurable parameter $Q$, which is the desired quantile of order $\alpha$ of the $\chi^2(3)$ distribution, where $\alpha$ is the confidence level of the Gaussian distribution. In practice, we compute the intersection between the line parallel to the ray and the ellipsoid formed by the points whose Mahalanobis distance to the mean of the Gaussian is equal to $Q$.

The computation of such an intersection is a well-defined problem that can be effectively solved using the following proposition (for the proof, see Appendix~\ref{app:raygaussint}).

\begin{proposition}[Following \cite{hearn2010computer}]\label{prop:1}
Consider the ray defined as ${\bf r}(t) = {\bf o} + t {\bf d}$, where ${\bf o}$ is the origin and ${\bf d}$ is the direction, and the Gaussian component $\N({\bf \mu}, \Sigma)$.
The first (closest to the origin ${\bf o}$) intersection between the ray ${\bf r}(t)$ and the confidence ellipse $\E_{ {\bf \mu}, \Sigma, q }$ is given by
\begin{equation}
    t_1 = \left \lbrace
\begin{array}{ll}
t_{*}, & \text{if } \left \langle {\bf o}', {\bf d}' \right \rangle \ge 0 \\
\frac{\left \langle {\bf o}' , {\bf o}' \right \rangle - Q}{\left \langle {\bf d}' , {\bf d}' \right \rangle  t_{*} }, & \text{if } \left \langle {\bf o}' {\bf d}' \right \rangle < 0
\end{array}
\right.,
\end{equation}
where
\begin{equation}
    \begin{array}{c}
t_{*} = - \frac{ \left \langle {\bf o}', {\bf d}' \right \rangle}{\left \langle {\bf d}', {\bf d}' \right \rangle} - \frac{\mathrm{sgn}\left \langle {\bf o}', {\bf d}' \right \rangle \sqrt{ Q - {\left \lVert {\bf o}' - \frac{{\bf d}'}{\left \lVert {\bf d}' \right \rVert} \left \langle {\bf o}', \frac{{\bf d}'}{ \| {\bf d}' \| } \right \rangle \right \rVert }^2  }}{\left \langle {\bf d}', {\bf d}' \right \rangle}.
\end{array}
\end{equation}

\end{proposition}

The application of the NVIDIA OptiX programming interface \cite{parker2010optix} and Proposition~\ref{prop:1} facilitates the efficient implementation of the detection of all ellipses possessing non-empty intersection with a ray in a numerically stable manner. Consequently, the Gaussian components aligned with the ray can be effectively determined in this phase.

\paragraph{Color Aggregation Along the Ray}

Suppose we have collected all the Gaussians along the ray defined as ${\bf r}(t) = {\bf o} + t {\bf d}$ (where ${\bf o}$ is the origin and ${\bf d}$ is the direction) in the following family:
\begin{equation}
    \G_{{\bf r}(t)} = \{(\N(\m_i,\Sigma_i), \hat{\alpha}_i, c_i) \}_{i=1}^{N}.
\end{equation}
Our methodology is similar to 3DGRT \cite{moenne20243d} in that we compute $\alpha_i$ by taking the product of the learned opacity for each Gaussian, denoted as $\hat{\alpha}_i$, and the peak of the 3D Gaussian probability density function scalar field normalized over the specific ray, i.e,

\begin{equation}\label{equ:ray}
\alpha_i = \hat{\alpha}_i \max\limits_{t \ge 0} \left\lbrace (2\pi)^{\frac{3}{2}} \sqrt{\left \lvert \Sigma_i \right \rvert} f_{\mathcal{N} \left( \m_i, \Sigma_i \right) } \left( {\bf r}(t) \right) \right \rbrace.
\end{equation}

\begin{figure}[t]
    \centering
        \begin{center}
        \end{center}
        \includegraphics[width=0.95\linewidth, trim=0 00 0 0 , clip]{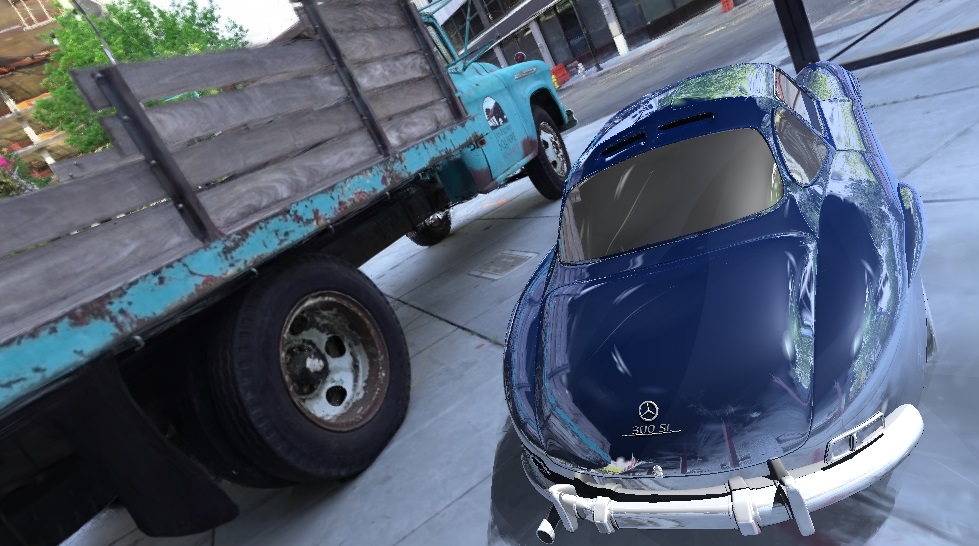} 
        \includegraphics[width=0.95\linewidth, trim=0 00 0 0 , clip]{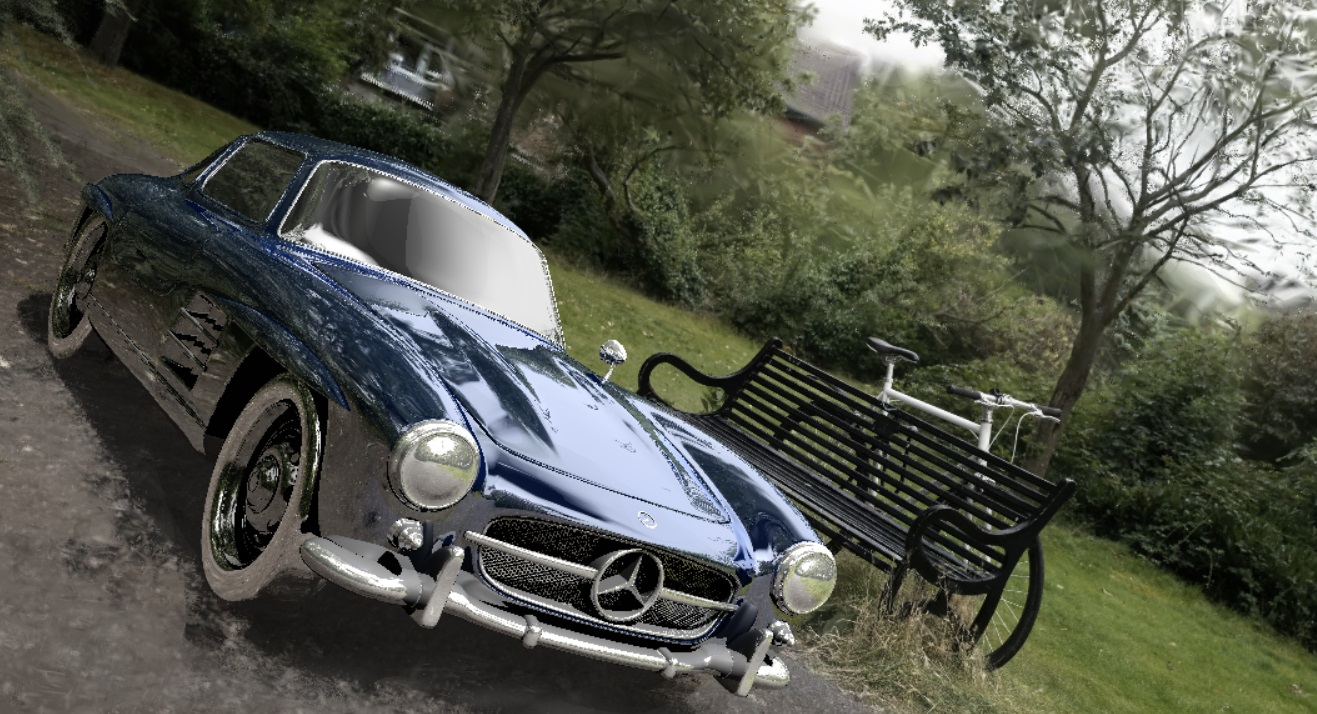} 
    \caption{\our{} (our) combines meshes with 3DGS-based representations with different material structures.}
    \label{fig:car}
\end{figure}

We can then calculate the Gaussian maximum response according to the 3DGRT \cite{moenne20243d}, using the following proposition (for the proof, see Appendix~\ref{app:alphai}).

\begin{proposition}[Following \cite{moenne20243d}]\label{prop:2}
Consider the ray defined as ${\bf r}(t) = {\bf o} + t {\bf d}$, where ${\bf o}$ is the origin and ${\bf d}$ is the direction, and a family of Gaussians $\G_{{\bf r}(t)} = \{ (\N(\m_i,\Sigma_i), \hat{\alpha}_i, c_i) \}_{i=1}^{N}$ that possess a non-empty intersection with the ray. Then we have
\begin{equation}
\begin{array}{cc}
\alpha_i \!\!\!\!\! & = \hat{\alpha}_i \max\limits_{t \ge 0} \left\lbrace (2\pi)^{\frac{3}{2}} \sqrt{\left \lvert \Sigma_i \right \rvert} f_{\mathcal{N} \left( \m_i, \Sigma_i \right) } \left( {\bf r}(t) \right) \right \rbrace \\
&= \hat{\alpha}_i \cdot e^{ -\frac{1}{2} \left \lVert {\bf o}' - \frac{{\bf d}'}{\left \lVert {\bf d}' \right \rVert} \left \langle \frac{{\bf d}'}{\left \lVert {\bf d}' \right \rVert }, {\bf o}' \right \rangle \right \rVert }
\end{array},
\end{equation}
where $\Sigma_i=R_i S_i S_i^T R_i^T$, ${\bf o}' = S_i^{-1} R_i^T \left( {\bf o} - \m_i\right)$, and ${\bf d}' = S_i^{-1} R_i^T {\bf d}$.

\end{proposition}


We emphasize that \our{} models $\alpha_i$ as the product of the trainable parameter $\hat{\alpha}_i$ and the value of the probability density function of the 3D Gaussian. Therefore, we use Eq.~\ref{equ:color} and the Gaussian maximum response for each component, as in Proposition~\ref{prop:2}.

In \our{}, a distinct color aggregation method is employed during the forward phase. In contrast to 3DGS and 3DGRT, which do not utilize the theoretical maximum for the number of intersectable Gaussians, imposing a threshold on the transmittance value $T_i$, \our{} applies a configurable upper bound on the Gaussians intersected by the ray in the forward phase. The enforcement of this limit enables the storage of the indices of all Gaussians intersected by the ray in an indices buffer, thereby preventing buffer overflow. This stored information is subsequently utilized in the backward phase for gradient computation, eliminating the need for re-traversing the rays. As ray traversal constitutes the most resource-intensive operation in ray tracing-based Multi-View Stereo tasks, this strategy reduces the number of ray-Gaussian intersections calculated by the algorithm by half.

In addition, \our{} introduces the second additional threshold for computing the gradient of the last ``meaningful'' Gaussian. If $T_i$ falls below the first threshold, we do not kill the ray immediately, but allow it to traverse the scene and hit some more Gaussians, since they carry the information necessary to compute the gradient for the last ``meaningful'' Gaussian, i.e., the Gaussian with index $i$ for which $T_i$ falls below the first threshold.

To illustrate the \our{} concept, consider the situation where the first Gaussian hit by the ray is close to fully opaque (i.e., $\alpha_1 \approx 1$). Hence $T_1 = (1 - \alpha_1) \approx 0$. Obviously, the subsequent Gaussians along the ray have no effect on the final aggregated color. Note, however, that the second component of the gradient for $\alpha_1$ of the aggregated color, which is given by the following formula:
\begin{equation}
    \dv{C}{{\alpha}_1} = c_1 - \sum \limits_{j=2}^N c_j {\alpha}_j \left( \prod \limits_{ \substack{ k \ne 1 } }^{j-1} \left( 1 - \alpha_k \right) \right),
\end{equation}
can be arbitrarily large, since the factor $1 - {\alpha}_1$ has been reduced. Therefore, we use:
\begin{equation}
   \prod_{k=i+1}^{j-1} \left( 1 - \alpha_k \right) < {\varepsilon}_2
\end{equation}
as a ray termination criterion before intersecting the $j$-th Gaussian in the final phase where the Gaussians are collected for the last ``meaningful'' Gaussian gradient calculation.


The algorithm below provides more details for the color aggregation operation in the forward phase.

Algorithm~1 provides more details for the color aggregation operation in the forward phase. As long as the actual number of Gaussians (the value of the variable $k$) does not exceed the configurable maximum allowed number of Gaussians (line 5), we traverse the ray (line 6), and if the intersection is not empty (line 7), we aggregate the output image color $I$ (lines ~11--~19). Otherwise, we set the sentinel value in the indices buffer to $-1$ for the sake of gradient computation in the backward phase (line 8) and terminate the algorithm (line 9). If, after the color aggregation, the transmittance $T_1$ falls below the first threshold, i.e., $\varepsilon_1$ (line 20), we set the variable second\_phase (line ~22), indicating that it is time to proceed to the second phase of the algorithm, where the Gaussians are collected for the sake of the last ``meaningful'' gradient computation, if it was not yet set, or update the transmittance $T_2$ otherwise (line ~24). Finally, we check (line ~27) if the transmittance $T_2$ falls below the second threshold, i.e. $\varepsilon_2$. If so, we terminate the algorithm after setting the sentinel value in the indices buffer (lines ~28--~29) or ``move'' the ray origin $\bf o$ a bit so that the ray does not hit the Gaussian with the index given by the index variable again during the next run of the for loop (line ~33).

\paragraph{Loss Function of the \our{} Model}
Given that the RGB color space is employed, the loss function is defined as the average of loss functions calculated independently for the specific RGB components.
We employ the analogical loss as in the 3DGS; however, we replace the $L_1$ norm with the $L_2$ norm and the spherical harmonics with the RGB color components. This results in the following formula for the loss function of each color component:
\begin{equation}
    \L= (1-\lambda)\ L_2 + \lambda \L_\text{D-SSIM},
\end{equation}
where $L_2$ is the squared error between the predicted and target color component values, ensuring color fidelity, $\L_\text{D-SSIM}$ is a structural similarity loss that preserves fine details and texture consistency in the rendered image, and $\lambda$ controls the balance between these two terms, allowing a trade-off between pixel-wise accuracy and perceptual similarity.

  \includegraphics[width=0.95\linewidth]{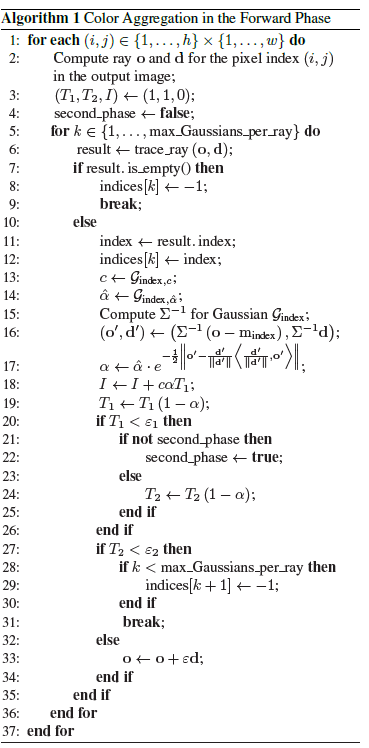}





\begin{figure*}[h]
    \centering
    \includegraphics[width=1\linewidth]{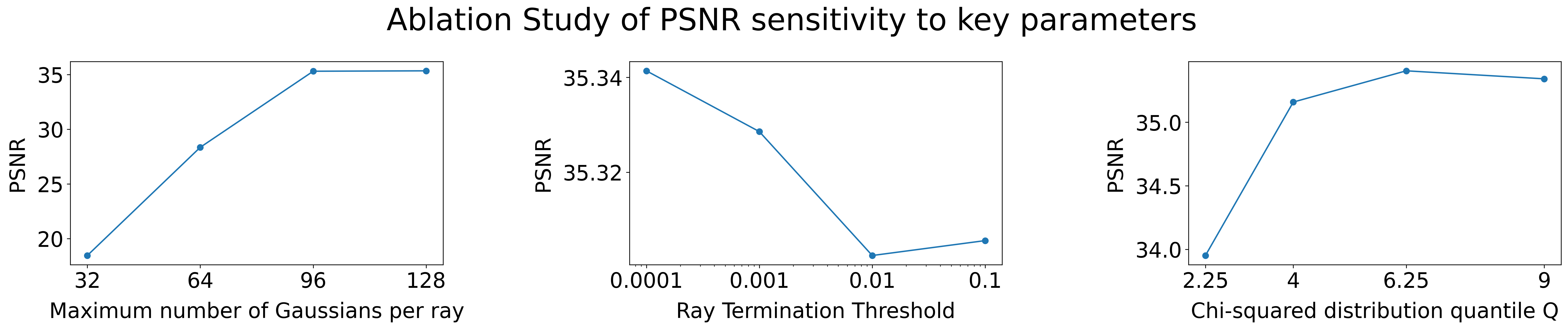}
    \caption{Ablation study investigating the effect of three key parameters of the \our{} model (our): the upper limit of Gaussians that can be hit by the ray, the ray termination threshold $\varepsilon_1$ used throughout the forward phase, and the quantile $Q$ of order $\alpha$ of the $\chi^2(3)$ distribution. The results are presented in terms of the PSNR  metric (greater is better).}
    \label{fig:ablation}
\end{figure*}



\section{Experiments}

In this section, we present the findings from our experimental study, encompassing both quantitative and qualitative results. In addition, we have conducted an ablation study to further explore the impact of selected key parameters of \our{}.

\begin{table*}[ht!]
       \caption{Quantitative evaluation of \our{} (our) on the Mip-NeRF360 \citep{barron2022mip}, Tanks and Temples \citep{knapitsch2017tanks}, and Deep Blending \citep{hedman2018deep} datasets. Comparison is made with the following baselines: Plenoxels \citep{fridovich2022plenoxels},
    INGP \citep{muller2022instant}, M-NeRF360 \citep{barron2021mip}, 3DGS \citep{kerbl20233d}, and 3DGRT \citep{moenne20243d}.\\[-5mm]
    }
{
\begin{center}
    \begin{tabular}{@{\;\;}c@{\;\;\;\;}c@{\;\;\;\;}c@{\;\;\;\;}c@{\;\;\;\;}c@{\;\;\;\;}c@{\;\;\;\;}c@{\;\;\;\;}c@{\;\;\;\;}c@{\;\;\;\;}c@{\;\;\;\;}c@{\;\;}}
    &
    &  \multicolumn{3}{c}{Mip-NeRF360} & \multicolumn{3}{c}{Tanks\&Temples} & \multicolumn{3}{c}{Deep Blending} \\
    \toprule
         & & SSIM $\uparrow$& PSNR $\uparrow$& LPIPS $\downarrow$& SSIM $\uparrow$& PSNR $\uparrow$& LPIPS $\downarrow$ & SSIM $\uparrow$& PSNR $\uparrow$& LPIPS $\downarrow$
 \\ \midrule
\multirow{7}{*}{\rotatebox{90}{ 
Spherical Harmonics
}}
&
Plenoxels & 0.626 & 23.08 &  0.719 & 0.379& 
21.08& 0.795 & 
0.510& 23.06& 0.510
\\
&
INGP-Base & 0.671 & 25.30 & 0.371 & 0.723 & 21.72 & 0.330 & 0.797 & 23.62 & 0.423
\\
&
INGP-Big & 0.699 & 25.59 & 0.331 & 0.745 & 21.92 & 0.305 & 0.817 & 24.96 & 0.390
\\
&
M-NeRF360 & \yellowc 0.792 & \redc 27.69 & \orangec 0.237 & 0.759 & \yellowc 22.22 & 0.257 & \orangec 0.901 & \yellowc 29.40 & \orangec 0.245
\\[0.001cm]

&
3DGS-7K &  0.770 &  25.60 &  0.279 & 0.767 & 21.20 & 0.280 & 0.875 & 27.78 & 0.317
\\
&
3DGS-30K & \orangec  0.815 & \yellowc 27.21 & \redc 0.214  & \redc 0.841 & \orangec 23.14 &  \redc 0.183 & \redc 0.903 & \orangec 29.41 & \redc 0.243\\
&
3DGRT &  - & - & -  &  \orangec 0.830 & \redc 23.20 &  \yellowc 0.222 & \yellowc 0.900 &  29.23 &  \yellowc 0.315\\
\midrule
RGB &
\our{} & \redc  0.846 & \orangec 27.31 & \orangec 0.237  &  \yellowc 0.829 &  22.20 & \orangec 0.202 &  \yellowc 0.900 & \redc 29.57 &  0.320\\
\bottomrule     
    \end{tabular}
    \end{center}
    }
    \label{tab:scene_mip_tt_db}
\end{table*} 
\paragraph{Datasets and Metrics}
Following \cite{moenne20243d} we conduct experiments on three widely used datasets: Mip-NeRF 360 \citep{barron2022mip}, Tanks and Temples \citep{knapitsch2017tanks}, and Deep Blending \citep{hedman2018deep}.
For consistency, we evaluate the same scenes as in \cite{moenne20243d}.
Specifically, for Mip-NeRF360, we use four indoor scenes: room, counter, kitchen and bonsai, as well as three outdoor scenes: bicycle, garden and stump.
On the Tanks and Temples dataset, we evaluate two large outdoor scenes: train and truck. For the Deep Blending dataset, we use two indoor scenes: playroom and drjohnson. 
In line with prior work, all evaluations use images downsampled by a factor of two for indoor scenes and by a factor of four for outdoor scenes.

We maintain consistent train/test splits across all datasets. For evaluation, we use three widely used metrics: PSNR, SSIM \cite{wang2004image}, and LPIPS \cite{zhang2018unreasonable}.

\paragraph{Quantitative Results}
In most cases, our\our{}  model demonstrates the capacity to attain outcomes that are analogous to those of classical 3D Gaussian Splatting, as illustrated in Table~\ref{tab:scene_mip_tt_db}. It is noteworthy that our approach employs RGB colors instead of spherical harmonics, a choice that aligns with the prevalent utilization of RGB color representation in conjunction with ray tracing. Furthermore, our model facilitates the incorporation of light effects, transparency, and shadows through the employment of ray tracing techniques. This approach eliminates the necessity of relying on formats contingent on viewing directions, thereby ensuring a more robust and consistent results.

\paragraph{Qualitative Results}

\begin{figure} 
    \centering
     \includegraphics[width=0.99\linewidth]{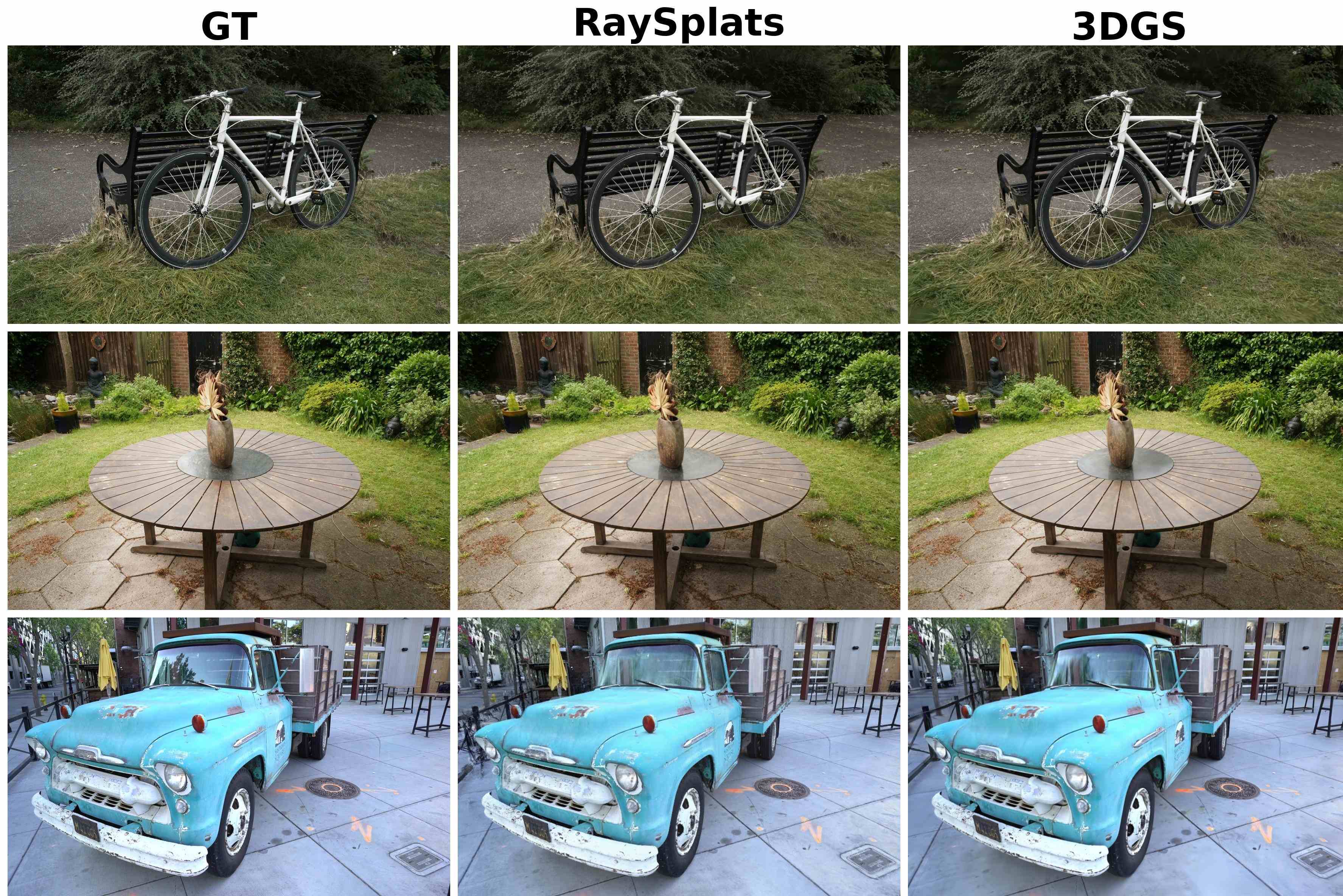}
    \caption{Examples of renderings of different types: the first column shows the ground truth image, the second column shows the rendering of the optimized \our{}, and the third column shows the rendering of 3D Gaussian splatting with RGB colors. The first two rows consist of data from the Mip-NeRF360 dataset, while the last row consists of the Truck example from the Tanks\&Templates dataset. Note that \our{} gives comparable results to the classic 3DGS.}
    \label{fig:big-scene_nvdiffrast}
\end{figure}

By leveraging ray tracing, which enables the simulation of lighting conditions, \our{} incorporates RGB colors. Nevertheless, in almost every practical scenario, our model delivers outcomes that are on par with traditional 3D Gaussian Splatting employing spherical harmonics.
Specifically, as illustrated in Fig.~\ref{fig:big-scene_nvdiffrast}, \our{} consistently delivers high-quality reconstructions. Additionally, it is compatible with mesh-based models that can undergo modifications through the incorporation of ray-tracing effects. As illustrated in Fig.~\ref{fig:3glasses}, the integration of glass properties within the model facilitates its connection to meshes. The incorporation of shadows, transparency, and light reflections serves to enhance the visual fidelity of the model. Fig.~\ref{fig:mirror} demonstrates model's capacity to mirror Gaussian reflections, while Fig.~\ref{fig:glass} showcases the glass dragon on a Lego structure from the NeRF synthetic dataset. Additionally, Fig.~\ref{fig:car} illustrates the versatility of our technique by showcasing it on a variety of colored meshes. It is evident that our approach is capable of representing both reflections and shadows with precision.

\paragraph{Ablation Study}
\our{} includes several key parameters that determine the number of Gaussians combined along the ray and have the impact on quality of the reconstructed 3D scene. In Fig.~\ref{fig:ablation} we present an ablation study of three such parameters: the parameter $Q$ being the quantile of order $\alpha$ of the $\chi^2(3)$-distribution (with three degrees of freedom) where $\alpha$ is the Gaussian distribution confidence level, the upper limit of Gaussians that can be hit by the ray, and the ray termination threshold $\varepsilon_1$ used throughout the forward phase.

Based on the results, while it is always profitable (from the perspective of the loss function) to increase the upper limit of Gaussians that can be hit by the ray (which seems to be in line with related techniques), increasing the value of $Q$ and decreasing $\varepsilon_1$ seems to pay off only up to a certain point where the value of the PSNR statistic begins to slowly decrease. During our experiments, we set these parameters (individually for each of the datasets) to values that represent a trade-off between reconstruction quality (in terms of PSNR value) and performance.

\section{Implementation Details}
We developed our technique in C/C++ using the NVIDIA OptiX 8.0.0 SDK. We implemented separate CUDA kernels for the forward color aggregation phase and the backward phase, where the gradient is computed based on the Gaussian indices stored in the index buffer. The entire gradient computation process was implemented manually from scratch, without recourse to the automatic differentiation systems, based on the formulas derived in Appendices \ref{app:grad1}, \ref{app:grad2}, \ref{app:app0}, \ref{app:grad3}, and \ref{app:grad4}. To compute the gradient of the loss function, we swept the successive indices of the Gaussians stored in the buffer and, after computing the value of $\dv{I}{{{}\Box{}}_i}$, we accumulated the derivative with respect to the ${}\Box{}$ parameter of the $i$-th Gaussian in the traversal order using the global memory atomicAdd() operation. Thanks to the reuse of the previously computed values of $\dv{I}{{\alpha}_i}$, we were able to compute each of the derivatives in $\mathcal{O} \left( 1 \right)$ time with respect to the number of stored Gaussian indices $N$, without having to compute the value of $\dv{I}{{\alpha}_i}$ from scratch in $\mathcal{O} \left( N \right)$ time. If the value of $\dv{I}{{\alpha}_i}$ was not finite or the transmittance $T_i$ fell below the threshold $\varepsilon_1$, we treated such a Gaussian as the last ``meaningful'' Gaussian (from the perspective of color aggregation) and computed the derivatives with respect to its parameters using the regular naive formula based on the Gaussians with indices stored in the successive remaining entries of the index buffer.

We optimized the parameters of our model using the ADAM optimizer (also implemented by hand) with the following values of the hyperparameters: $\beta_1 = 0.9$, $\beta_2 = 0.999$, and $\varepsilon = 0.00000001$. For the Gaussian parameters $\hat{\alpha}$, $s_x$, $s_y$, and $s_z$ (according to \cite{kerbl20233d}) we used the inverse sigmoid function. Our technique is based on the similar densification strategy as in \cite{kerbl20233d} except for the parameters $m_x$, $m_y$, and $m_z$, since we do not use the norm of the positional gradient but the ``regular'' geometric gradient in the densification criterion. Note that the former does not make sense in the more advanced ray tracing settings that we intend to explore in our future work. In our technique, we also do not periodically set the Gaussian trainable opacities $\hat{\alpha}$ to the value close to $0$, as this may lead to a decrease in the precision of the loss function gradient. This is due to the configurable upper limit of Gaussians the ray can hit during the traversing phase.

\section{Conclusions}

In this paper, we introduced \our{}, a novel approach that integrates ray tracing into 3D Gaussian Splatting (3DGS) to overcome the limitations of traditional rasterization-based methods. By directly operating on Gaussian primitives represented by confidence ellipses with RGB colors, our method enables more accurate light and shadow interactions. The intersection of ellipses and rays is computed to construct a ray-tracing framework that enhances 3DGS with realistic lighting effects. In consequence, the incorporation of meshes, lights, and shadows is facilitated, resulting in a significant enhancement of the visual fidelity of 3D Gaussian Splatting models.

\paragraph{Limitations}
Our method requires a specialized renderer that supports both ray tracing and 3D Gaussian Splatting.

\section*{Impact Statement}

This paper presents work that aims to advance the field of Machine Learning. There are many potential societal consequences of our work, none which we feel must be specifically highlighted here.

\nocite{langley00}

\bibliographystyle{icml2024}


\newpage
\appendix
\onecolumn

\section{Theoretical Analysis}
\label{appendix}

\subsection{Computing a Ray-Gaussian Intersection}\label{app:raygaussint}

As explained in Section \ref{intersection}, the ray-Gaussian intersection is the intersection between the ray and the Gaussian confidence ellipsoid parameterized by the value of the configurable parameter $Q$, which is the desired quantile of order $\alpha$ of the $\chi^2(3)$ distribution, where $\alpha$ is the Gaussian confidence level. Obviously, the points ${\bf x} \in \mathbb{R}^3$ belonging to the above ellipsoid satisfy the following equation:
$$
\left({\bf x} - \m_i \right)^T \Sigma_i^{-1} \left({\bf x} - \m_i \right) = Q
$$
Substituting the ray parametric equation in the $t$ variable into the above equation yields:
$$
\left({\bf r}(t) - \m_i \right)^T \Sigma_i^{-1} \left({\bf r}(t) - \m_i \right) = Q \iff
\left({\bf r}(t) - \m_i \right)^T \Sigma_i^{-1} \left({\bf r}(t) - \m_i \right) - Q = 0
$$
$\Sigma_i=R_i S_i S_i^T R_i^T = R_i S_i S_i R_i^T$, therefore:

$$\Sigma_i^{-1} = R_i S_i^{-1} S_i^{-1} R_i^T$$
and
\begin{flalign*}
& \left({\bf r}(t) - \m_i \right)^T R_i S_i^{-1} S_i^{-1} R_i^T \left({\bf r}(t) - \m_i \right) - Q = 0 \iff &\\
& \iff {\left( S_i^{-1} R_i^T \left( {\bf r}(t) - \m_i \right) \right)}^T \left( S_i^{-1} R_i^T \left( {\bf r}(t) - \m_i \right) \right) - Q = 0 \iff &\\
& \iff {\left( S_i^{-1} R_i^T \left( {\bf o} + t{\bf d} - \m_i \right) \right)}^T \left( S_i^{-1} R_i^T \left( {\bf o} + t{\bf d} - \m_i \right) \right) - Q = 0 \iff &\\
& \iff {\left( S_i^{-1} R_i^T \left( \left( {\bf o} - \m_i \right) + t{\bf d} - \vec{0} \right) \right)}^T \left( S_i^{-1} R_i^T \left( \left( {\bf o} - \m_i \right) + t{\bf d} - \vec{0} \right) \right) - Q = 0
\end{flalign*}
Let us define ${\bf o}' \coloneq S_i^{-1} R_i^T \left( {\bf o} - \m_i \right)$ and ${\bf d}' \coloneq S_i^{-1} R_i^T {\bf d}$. Then:
\begin{flalign*}
& {\left( S_i^{-1} R_i^T \left( \left( {\bf o} - \m_i \right) + t{\bf d} - \vec{0} \right) \right)}^T \left( S_i^{-1} R_i^T \left( \left( {\bf o} - \m_i \right) + t{\bf d} - \vec{0} \right) \right) - Q = 0 \iff &\\
& \iff { \left( {\bf o}'+ t{\bf d}' - \vec{0} \right) }^T \left( {\bf o}'+ t{\bf d}' - \vec{0} \right) - Q = 0
\end{flalign*}
So we transformed the problem of the ray-Gaussian intersection into the problem of the ray-sphere intersection for the sphere with the center at the origin of the coordinate system.
Now:
\begin{flalign*}
& { \left( {\bf o}'+ t{\bf d}' - \vec{0} \right) }^T \left( {\bf o}'+ t{\bf d}' - \vec{0} \right) - Q = 0 \iff &\\
& \iff { \left( {\bf o}'+ t{\bf d}' \right) }^T \left( {\bf o}'+ t{\bf d}' \right) - Q = 0 \iff &\\
& \iff \left \langle {\bf o}' , {\bf o}' \right \rangle + 2 \left \langle {\bf o}', {\bf d}' \right \rangle \left( t \right) + \left \langle {\bf d}', {\bf d}' \right \rangle \left( t^2 \right) - Q = 0 \iff &\\
& \iff \left \langle {\bf d}', {\bf d}' \right \rangle \left( t^2 \right) + 2 \left \langle {\bf o}', {\bf d}' \right \rangle \left( t \right) + \left( \left \langle {\bf o}' , {\bf o}' \right \rangle - Q \right) = 0
\end{flalign*}
Let us define: $a \coloneq \left \langle {\bf d}', {\bf d}' \right \rangle$, $b \coloneq 2 \left \langle {\bf o}', {\bf d}' \right \rangle$ and $c \coloneq \left \langle {\bf o}' , {\bf o}' \right \rangle - Q$. 
Since the classical way of computing the $\Delta$ to solve the above quadratic equation suffers from numerical stability problems, we will show, according to [Hearn and Baker], how to compute it in a more numerically robust way.
\begin{flalign*}
\Delta &= b^2-4a{c} =&\\
&= 4a \left( \frac{b^2}{4a} - c \right) =&\\
&= 4 \left \langle {\bf d}', {\bf d}' \right \rangle \left( \frac{\bcancel{4} {\left \langle {\bf o}', {\bf d}' \right \rangle}^2 }{\bcancel{4} \left \langle {\bf d}', {\bf d}' \right \rangle} - \left( \left \langle {\bf o}' , {\bf o}' \right \rangle - Q \right) \right) =&\\
&= 4 \left \langle {\bf d}', {\bf d}' \right \rangle \left( \frac{ {\left \langle {\bf o}', {\bf d}' \right \rangle}^2 }{ \left \langle {\bf d}', {\bf d}' \right \rangle} - \left( \left \langle {\bf o}' , {\bf o}' \right \rangle - Q \right) \right) =&\\
&= 4 \left \langle {\bf d}', {\bf d}' \right \rangle \left( Q - \left( \left \langle {\bf o}' , {\bf o}' \right \rangle - \frac{ {\left \langle {\bf o}', {\bf d}' \right \rangle}^2 }{ \left \langle {\bf d}', {\bf d}' \right \rangle} \right) \right) =&\\
&= 4 \left \langle {\bf d}', {\bf d}' \right \rangle \left( Q - \left( \left \langle {\bf o}' , {\bf o}' \right \rangle - \frac{1}{{\left \lVert {\bf d}' \right \rVert}^2} {\left \langle {\bf o}', {\bf d}' \right \rangle}^2 \right) \right) =&\\
&= 4 \left \langle {\bf d}', {\bf d}' \right \rangle \left( Q - \left( { \left \langle {\bf o}' , {\bf o}' \right \rangle - \left( \frac{1}{ \left \lVert {\bf d}' \right \rVert } \left \langle {\bf o}', {\bf d}' \right \rangle \right) }^2 \right) \right) =&\\
&= 4 \left \langle {\bf d}', {\bf d}' \right \rangle \left( Q - \left( \left \langle {\bf o}' , {\bf o}' \right \rangle - {\left \langle {\bf o}', \frac{{\bf d}'}{\left \lVert {\bf d}' \right \rVert} \right \rangle}^2 \right) \right)
\end{flalign*}
Using the Pythagorean theorem and some vector arithmetic, we finally get the following:
$$
\Delta =
4 \left \langle {\bf d}', {\bf d}' \right \rangle \left( Q - {\left \lVert {\bf o}' - \frac{{\bf d}'}{\left \lVert {\bf d}' \right \rVert} \left \langle {\bf o}', \frac{{\bf d}'}{\left \lVert {\bf d}' \right \rVert} \right \rangle \right \rVert}^2 \right)
$$
Hence:
\begin{flalign*}
t_{1{2}} &=
\frac{-b \mp \sqrt{\Delta}}{2a} =&\\
&= \frac{-2 \left \langle {\bf o}', {\bf d}' \right \rangle \mp \sqrt{4 \left \langle {\bf d}', {\bf d}' \right \rangle \left( Q - {\left \lVert {\bf o}' - \frac{{\bf d}'}{\left \lVert {\bf d}' \right \rVert} \left \langle {\bf o}', \frac{{\bf d}'}{\left \lVert {\bf d}' \right \rVert} \right \rangle \right \rVert}^2 \right)}}{2 \left \langle {\bf d}', {\bf d}' \right \rangle} =&\\
&= \frac{-\bcancel{2} \left \langle {\bf o}', {\bf d}' \right \rangle \mp \bcancel{2} \sqrt{ \left \langle {\bf d}', {\bf d}' \right \rangle \left( Q - {\left \lVert {\bf o}' - \frac{{\bf d}'}{\left \lVert {\bf d}' \right \rVert} \left \langle {\bf o}', \frac{{\bf d}'}{\left \lVert {\bf d}' \right \rVert} \right \rangle \right \rVert}^2 \right)}}{\bcancel{2} \left \langle {\bf d}', {\bf d}' \right \rangle} =&\\
&= \frac{- \left \langle {\bf o}', {\bf d}' \right \rangle \mp \sqrt{ \left \langle {\bf d}', {\bf d}' \right \rangle \left( Q - {\left \lVert {\bf o}' - \frac{{\bf d}'}{\left \lVert {\bf d}' \right \rVert} \left \langle {\bf o}', \frac{{\bf d}'}{\left \lVert {\bf d}' \right \rVert} \right \rangle \right \rVert}^2 \right)}}{ \left \langle {\bf d}', {\bf d}' \right \rangle}
\end{flalign*}
Since naive roots can be numerically unstable when
$$
- \left \langle {\bf o}', {\bf d}' \right \rangle \approx \sqrt{ \left \langle {\bf d}', {\bf d}' \right \rangle \left( Q - {\left \lVert {\bf o}' - \frac{{\bf d}'}{\left \lVert {\bf d}' \right \rVert} \left \langle {\bf o}', \frac{{\bf d}'}{\left \lVert {\bf d}' \right \rVert} \right \rangle \right \rVert}^2 \right)},
$$
we obtain the desired $t$ parameter value for the nearest intersection point using the Viete formula, in case it would lead to numerical instability:
$$
t_{*} = \frac{- \left \langle {\bf o}', {\bf d}' \right \rangle - \mathrm{sgn}\left( \left \langle {\bf o}', {\bf d}' \right \rangle \right) \sqrt{ \left \langle {\bf d}', {\bf d}' \right \rangle \left( Q - {\left \lVert {\bf o}' - \frac{{\bf d}'}{\left \lVert {\bf d}' \right \rVert} \left \langle {\bf o}', \frac{{\bf d}'}{\left \lVert {\bf d}' \right \rVert} \right \rangle \right \rVert}^2 \right) } }{ \left \langle {\bf d}', {\bf d}' \right \rangle}
$$
$$
t_1 = \left \lbrace
\begin{array}{lr}
t_{*} & , \left \langle {\bf o}', {\bf d}' \right \rangle \ge 0 \\
\frac{\left \langle {\bf o}' , {\bf o}' \right \rangle - Q}{\left \langle {\bf d}' , {\bf d}' \right \rangle t_{*} } & , \left \langle {\bf o}', {\bf d}' \right \rangle < 0 \\
\end{array}
\right.
$$

\subsection{Computing $\alpha_i$ Based on the Learned Per-Gaussian Opacity $\hat{\alpha}_i$}\label{app:alphai}

$$
\alpha_i =
\hat{\alpha}_i \max\limits_{t \ge 0} \left\lbrace (2\pi)^{\frac{3}{2}} \sqrt{\left \lvert \Sigma_i \right \rvert} f_{\mathcal{N} \left( \m_i, \Sigma_i \right) } \left( {\bf r}(t) \right) \right \rbrace =
\hat{\alpha}_i \max\limits_{t \ge 0} \left\lbrace e^{-\frac{1}{2}{\left({\bf r}(t)-\m_i \right)}^T \Sigma_i^{-1} \left({\bf r}(t)-\m_i \right) } \right \rbrace
$$

Since $e^x$ is the increasing function, we obtain:

\begin{flalign*}
\max \limits_{t \ge 0} \left\lbrace e^{-\frac{1}{2}{\left({\bf r}(t)-\m_i \right)}^T \Sigma_i^{-1} \left({\bf r}(t)-\m_i \right) } \right \rbrace &=
e^{\max \limits_{t \ge 0} \left\lbrace -\frac{1}{2}{\left({\bf r}(t)-\m_i \right)}^T \Sigma_i^{-1} \left({\bf r}(t)-\m_i \right) \right \rbrace} =&\\
&= e^{- \min \limits_{t \ge 0} \left\lbrace { \frac{1}{2} \left({\bf r}(t)-\m_i \right)}^T \Sigma_i^{-1} \left({\bf r}(t)-\m_i \right) \right \rbrace} =&\\
&= e^{-\frac{1}{2} \min \limits_{t \ge 0} \left\lbrace { \left({\bf r}(t)-\m_i \right)}^T \Sigma_i^{-1} \left({\bf r}(t)-\m_i \right) \right \rbrace}
\end{flalign*}
$\Sigma_i=R_i S_i S_i^T R_i^T = R_i S_i S_i R_i^T$, therefore:

$$\Sigma_i^{-1} = R_i S_i^{-1} S_i^{-1} R_i^T$$
and

\begin{flalign*}
\min \limits_{t \ge 0} \left\lbrace {\left({\bf r}(t)-\m_i \right)}^T \Sigma_i^{-1} \left({\bf r}(t)-\m_i \right) \right \rbrace &=
\min \limits_{t \ge 0} \left\lbrace {\left({\bf r}(t)-\m_i \right)}^T R_i S_i^{-1} S_i^{-1} R_i^T \left({\bf r}(t)-\m_i \right) \right \rbrace =&\\
&= \min \limits_{t \ge 0} \left \lbrace { \left( S_i^{-1} R_i^T \left({\bf r}(t)-\m_i \right) \right) }^T \left( S_i^{-1} R_i^T \left({\bf r}(t)-\m_i \right) \right) \right \rbrace =&\\
&= \min \limits_{t \ge 0} \left \lbrace { \left( S_i^{-1} R_i^T \left({\bf o} + t{\bf d} - \m_i \right) \right) }^T \left( S_i^{-1} R_i^T \left({\bf o} + t{\bf d} - \m_i \right) \right) \right \rbrace =&\\
&= \min \limits_{t \ge 0} \left \lbrace { \left( S_i^{-1} R_i^T \left(t{\bf d} + \left( {\bf o} - \m_i \right) \right) \right) }^T \left( S_i^{-1} R_i^T \left(t{\bf d} + \left( {\bf o} - \m_i \right) \right) \right) \right \rbrace
\end{flalign*}

Let us define: ${\bf o}' \coloneq S_i^{-1} R_i^T \left( {\bf o} - \m_i \right)$ and ${\bf d}' \coloneq S_i^{-1} R_i^T {\bf d}$. Then:

\begin{flalign*}
\min \limits_{t \ge 0} \left\lbrace {\left({\bf r}(t)-\m_i \right)}^T \Sigma_i^{-1} \left({\bf r}(t)-\m_i \right) \right \rbrace &=
\min \limits_{t \ge 0} \left \lbrace { \left( S_i^{-1} R_i^T \left(t{\bf d} + \left( {\bf o} - \m_i \right) \right) \right) }^T \left( S_i^{-1} R_i^T \left(t{\bf d} + \left( {\bf o} - \m_i \right) \right) \right) \right \rbrace =&\\
&= \min \limits_{t \ge 0} \left \lbrace { \left( S_i^{-1} R_i^T \left(t{\bf d}' + {\bf o}' \right) \right) }^T \left( S_i^{-1} R_i^T \left(t{\bf d}' + {\bf o}' \right) \right) \right \rbrace =&\\
&= \min \limits_{t \ge 0} \left \lbrace \left \lVert t{\bf d}' + {\bf o}' \right \rVert \right \rbrace =&\\
&= \min \limits_{t \ge 0} \left \lbrace \left \lVert t{\bf d}' - {\bf o}' \right \rVert \right \rbrace,
\end{flalign*}
where the last equality holds because the distance between the line and the point in $\mathbb{R}^3$ is exactly the same as the distance between the line and the symmetry of that point about the origin of the coordinate system. So we conclude that $t$ minimizes the Euclidean distance between $t {\bf d}'$ and ${\bf o}'$, and the exact minimum is the Euclidean distance between the line and the point ${\bf o}'$, given by the following formula:
$$
\min \limits_{t \ge 0} \left \lbrace \left \lVert t{\bf d}' - {\bf o}' \right \rVert \right \rbrace =
\left \lVert {\bf o}' - \frac{{\bf d}'}{\left \lVert {\bf d}' \right \rVert} \left \langle \frac{{\bf d}'}{\left \lVert {\bf d}' \right \rVert }, {\bf o}' \right \rangle \right \rVert
$$
Thus:
\begin{flalign*}
\alpha_i &=
\hat{\alpha}_i \max\limits_{t \ge 0} \left\lbrace e^{-\frac{1}{2}{\left({\bf r}(t)-\m_i \right)}^T \Sigma_i^{-1} \left({\bf r}(t)-\m_i \right) } \right \rbrace =&\\
&= \hat{\alpha}_i \cdot e^{ \frac{1}{2} \min \limits_{t \ge 0} \left\lbrace {\left({\bf r}(t)-\m_i \right)}^T \Sigma_i^{-1} \left({\bf r}(t)-\m_i \right) \right \rbrace} =&\\
&= \hat{\alpha}_i \cdot e^{ -\frac{1}{2} \min \limits_{t \ge 0} \left \lbrace \left \lVert t{\bf d}' - {\bf o}' \right \rVert \right \rbrace } =&\\
&= \hat{\alpha}_i \cdot e^{ -\frac{1}{2} \left \lVert {\bf o}' - \frac{{\bf d}'}{\left \lVert {\bf d}' \right \rVert} \left \langle \frac{{\bf d}'}{\left \lVert {\bf d}' \right \rVert }, {\bf o}' \right \rangle \right \rVert }
\end{flalign*}

\subsection{Computing $\dv{\L}{{}\Box{}}$ for Some Trainable Parameter ${}\Box{}$.}\label{app:grad1}
Recall that since we are using the RGB color space, the loss function $\L$ will be the average of the loss functions computed separately for the R, G, and B components of the successive pixels of both the rendered and the reference image:
$$
\L = \frac{1}{3}\left( \L_R + \L_G + \L_B \right)
$$
Thus, the gradient is given by the following formula:
$$
\dv{\L}{{}\Box{}} = \frac{1}{3}\left( \dv{\L_R}{{}\Box{}} + \dv{\L_G}{{}\Box{}} + \dv{\L_B}{{}\Box{}} \right)
$$
Considering that each of the loss function components is of the form:
$$
\L_{{}*{}} = (1-\lambda)\L_2^{{}*{}} + \lambda \L_\text{D-SSIM}^{{}*{}},
$$
the gradient $\dv{\L_{{}*{}}}{{}\Box{}}$ can be calculated as follows:
\begin{flalign*}
\dv{\L_{{}*{}}}{{}\Box{}} &=
\left( 1-\lambda \right) \dv{\L_2^{{}*{}}}{{}\Box{}} + \lambda \dv{\L_\text{D-SSIM}^{{}*{}}}{{}\Box{}} =&\\
&= \left( 1-\lambda \right) \dv{}{{}\Box{}} \left( \frac{1}{w \cdot h} \sum \limits_{i,j=1}^{h,w} { \left( I_{i,j}^{{}*{}} - {\hat{I}}_{i,j}^{{}*{}} \right) }^2 \right) + \lambda \dv{}{{}\Box{}} \left( \frac{1 - \L_{SSIM}^{{}*{}}}{2} \right) =&\\
&= \left( 1-\lambda \right) \left( \frac{1}{w \cdot h} \sum \limits_{i,j=1}^{h,w} \dv{}{{}\Box{}} { \left( I_{i,j}^{{}*{}} - {\hat{I}}_{i,j}^{{}*{}} \right) }^2 \right) - \frac{\lambda}{2} \cdot \dv{\L_{SSIM}^{{}*{}}}{{}\Box{}} =&\\
&= \left( 1-\lambda \right) \left( \frac{1}{w \cdot h} \sum \limits_{i,j=1}^{h,w} 2 \left( I_{i,j}^{{}*{}} - {\hat{I}}_{i,j}^{{}*{}} \right) \dv{I_{i,j}^{{}*{}}}{{}\Box{}} \right) - \frac{\lambda}{2} \left( \sum \limits_{i,j=1}^{h,w} \dv{\L_{SSIM}^{{}*{}}}{I_{i,j}^{{}*{}}} \cdot \dv{I_{i,j}^{{}*{}}}{{}\Box{}} \right) =&\\
&= \left( \sum \limits_{i,j=1}^{h,w} \frac{ 2 \left( 1-\lambda \right) }{w \cdot h} \left( I_{i,j}^{{}*{}} - {\hat{I}}_{i,j}^{{}*{}} \right) \right) \dv{I_{i,j}^{{}*{}}}{{}\Box{}} - \left( \sum \limits_{i,j=1}^{h,w} \frac{\lambda}{2} \cdot \dv{\L_{SSIM}^{{}*{}}}{I_{i,j}^{{}*{}}} \right) \dv{I_{i,j}^{{}*{}}}{{}\Box{}} =&\\
&= \sum \limits_{i,j=1}^{h,w} \left( \frac{ 2 \left( 1-\lambda \right) }{w \cdot h} \left( I_{i,j}^{{}*{}} - {\hat{I}}_{i,j}^{{}*{}} \right) - \frac{\lambda}{2} \cdot \dv{\L_{SSIM}^{{}*{}}}{I_{i,j}^{{}*{}}} \right) \dv{I_{i,j}^{{}*{}}}{{}\Box{}},
\end{flalign*}
where $h$ and $w$ are the image height and width, respectively (they have the same value for the rendered and the reference image), while $I_{i,j}^{{}*{}}$ and $\hat{I}_{i,j}^{{}*{}}$ are the color intensities at the point $\left( i,j \right)$ for the rendered and the reference image, respectively. For brevity in the later derivation, we will deliberately omit the indices: $i$ and $j$ in the color intensity formula, as well as the asterisk to indicate that the formula will be identical for all color components: R, G and B. As:
$$
I = \sum\limits_{i \in \N} c_i \alpha_i \left( \prod_{j=1}^{i-1} (1-\alpha_j) \right),
$$
where $\N$ is the set of indices \cite{kopanas2021point,kopanas2022neural} of Gaussians that intersect the ray after reordering the set of Gaussians so that the Gaussians with smaller indices are those hit "earlier" by the ray and the Gaussians with indices greater than $N$ are those having empty intersection, we obviously have:
$$
\dv{I}{{{}\Box{}}_i} = \left \lbrace
\begin{array}{lr}
\alpha_i \left( \prod \limits_{j=1}^{i-1} (1-\alpha_j) \right) & , \left( {}\Box{} = c \right) \land \left( i \in \N \right) \\
0 & , \left( {}\Box{} = c \right) \land \left( i \notin \N \right) \\
\dv{I}{\alpha_i} \cdot \dv{\alpha_i}{{{}\Box{}}_i} & , \left( {}\Box{} \ne c \right) \land \left( i \in \N \right) \\
0 & , \left( {}\Box{} \ne c \right) \land \left( i \notin \N \right)
\end{array}
\right.
$$
Now it remains to show how to compute $\dv{I}{\alpha_i}$, $\dv{\alpha_i}{{{}\Box{}}_i}$ and $\dv{\L_{SSIM}}{I_{i,j}}$.

\subsection{Computing $\dv{I}{\alpha_1}$ Using $I$.}\label{app:grad2}
Since
$$
I =
\sum\limits_{i=1}^N c_i{\alpha_i} \left( \prod\limits_{j=1}^{i-1} \left( 1-\alpha_j \right) \right),
$$
we have:
$$
\dv{I}{\alpha_1} =
c_1 - \sum\limits_{i=2}^N c_i{\alpha_i} \left( \prod\limits_{\substack{j=1 \\ j \ne 1 }}^{i-1} \left( 1-\alpha_j \right) \right)
$$
Let us multiply both sides of the equation by $1-\alpha_1$. Then:
\begin{flalign*}
\dv{I}{\alpha_1}\left( 1-\alpha_1 \right) &= c_1 \left( 1-\alpha_1 \right) - \sum\limits_{i=2}^N c_i{\alpha_i} \left( \prod\limits_{j=1}^{i-1} \left( 1-\alpha_j \right) \right) =&\\
&= c_1 \left( 1-\alpha_1 \right) - \left( \left( c_1{\alpha_1} - c_1{\alpha_1} \right) + \sum\limits_{i=2}^N c_i{\alpha_i} \left( \prod\limits_{j=1}^{i-1} \left( 1-\alpha_j \right) \right) \right) =&\\
&= c_1 \left( \left( 1 - \alpha_1 \right) + \alpha_1 \right) - \sum\limits_{i=1}^N c_i{\alpha_i} \left( \prod\limits_{j=1}^{i-1} \left( 1-\alpha_j \right) \right) =&\\
&= c_1 - I
\end{flalign*}
Hence:
$$
\dv{I}{\alpha_1} = \frac{c_1 - I}{1-\alpha_1},
$$
as long as $\alpha_1 \ne 1$. If $\alpha_1 = 1$ we use the ``regular'' formula.

\subsection{Computing $\dv{I}{\alpha_i}$ Using the Previous Value of $\dv{I}{\alpha_{i-1}}$ for $i > 1$.}\label{app:app0}
Since:
$$
\dv{I}{\alpha_i} =
c_i \left( \prod\limits_{j=1}^{i-1} \left( 1-\alpha_j \right) \right) - \sum\limits_{j=i+1}^N c_j{\alpha_j} \left( \prod\limits_{\substack{k=1 \\ k \ne i }}^{j-1} \left( 1-\alpha_k \right) \right)
$$
after multiplying both sides of the equation by $1-\alpha_i$ we obtain:
\begin{flalign*}
\dv{I}{\alpha_i}\left( 1-\alpha_i \right) &= c_i \left( \prod\limits_{j=1}^{i-1} \left( 1-\alpha_j \right) \right)\left( 1-\alpha_i \right) - \sum\limits_{j=i+1}^N c_j{\alpha_j} \left( \prod\limits_{k=1}^{j-1} \left( 1-\alpha_k \right) \right) =&\\
 &= c_i \left( \prod\limits_{j=1}^{i-1} \left( 1-\alpha_j \right) \right) \left( 1-\alpha_i \right) + \left( c_i{\alpha_i} - c_i{\alpha_i} \right)\left( \prod\limits_{j=1}^{i-1} \left( 1-\alpha_j \right) \right) - \sum\limits_{j=i+1}^N c_j{\alpha_j} \left( \prod\limits_{k=1}^{j-1} \left( 1-\alpha_k \right) \right) =&\\
 &= c_i \left( \prod\limits_{j=1}^{i-1} \left( 1-\alpha_j \right) \right) \left( \left( 1-\alpha_i \right) + \alpha_i \right) - \sum\limits_{j=i}^N c_j{\alpha_j} \left( \prod\limits_{k=1}^{j-1} \left( 1-\alpha_k \right) \right) =&\\
 &= c_i \left( \prod\limits_{j=1}^{i-1} \left( 1-\alpha_j \right) \right) - \sum\limits_{j=i}^N c_j{\alpha_j} \left( \prod\limits_{k=1}^{j-1} \left( 1-\alpha_k \right) \right) =&\\
 &= c_i \left( \prod\limits_{j=1}^{i-1} \left( 1-\alpha_j \right) \right) + \left( c_{i-1} - c_{i-1} \right) \left( \prod\limits_{j=1}^{i-1} \left( 1-\alpha_j \right) \right) - \sum\limits_{j=i}^N c_j{\alpha_j} \left( \prod\limits_{k=1}^{j-1} \left( 1-\alpha_k \right) \right) =&\\
 &= \left( c_i - c_{i-1} \right) \left( \prod\limits_{j=1}^{i-1} \left( 1-\alpha_j \right) \right) + c_{i-1} \left( \prod\limits_{j=1}^{i-1} \left( 1-\alpha_j \right) \right) - \sum\limits_{j=i}^N c_j{\alpha_j} \left( \prod\limits_{k=1}^{j-1} \left( 1-\alpha_k \right) \right) =&\\
 &= \left( c_i - c_{i-1} \right) \left( \prod\limits_{j=1}^{i-1} \left( 1-\alpha_j \right) \right) + \left( 1 - \alpha_{i-1} \right) \left( c_{i-1} \left( \prod \limits_{j=1}^{i-2} \left( 1-\alpha_j \right) \right) - \sum\limits_{j=i}^N c_j{\alpha_j} \left( \prod\limits_{\substack{k=1 \\ k \ne i-1}}^{j-1} \left( 1-\alpha_k \right) \right) \right) =&\\
 &= \left( c_i - c_{i-1} \right) \left( \prod\limits_{j=1}^{i-1} \left( 1-\alpha_j \right) \right) + \left( 1 - \alpha_{i-1} \right) \dv{I}{\alpha_{i-1}} &\\
 \end{flalign*}
Hence:
$$
\dv{I}{\alpha_i} = \frac{\dv{I}{\alpha_{i-1}} \left( 1 - \alpha_{i-1} \right) + \left( c_i - c_{i-1} \right) \left( \prod\limits_{j=1}^{i-1} \left( 1-\alpha_j \right) \right)}{1-\alpha_i}
$$
as long as $\alpha_i \ne 1$. If $\alpha_i = 1$ we use the ``regular'' formula. The quantities $\dv{I}{\alpha_{i-1}} \left( 1 - \alpha_{i-1} \right)$ and $\prod\limits_{j=1}^{i-1} \left( 1-\alpha_j \right)$ can easily be accumulated in two separate variables and updated each time the ray encounters the new Gaussian in the backward phase. Note, however, that the computational complexity is still linear in the case where $\alpha_i = 1$, since the Gaussians with indices greater than $i$ are not rendered because their transmittance $T_i$ is $0$.

\subsection{Computing $\dv{\alpha_i}{{{}\Box{}}_i}$.}\label{app:grad3}
Recall that:
$$
\alpha_i = \frac{1}{1 + e^{-\hat{\alpha}_i}} \cdot e^{ -\frac{1}{2} \left \lVert {\bf o}' - \frac{{\bf d}'}{\left \lVert {\bf d}' \right \rVert} \left \langle \frac{{\bf d}'}{\left \lVert {\bf d}' \right \rVert }, {\bf o}' \right \rangle \right \rVert }
$$
Therefore, we can immediately compute:
\begin{flalign*}
\dv{\alpha_i}{\hat{\alpha}_i} &=
\dv{}{\hat{\alpha}_i} \left( \frac{1}{1 + e^{-\hat{\alpha}_i}} \cdot e^{ -\frac{1}{2} \left \lVert {\bf o}' - \frac{{\bf d}'}{\left \lVert {\bf d}' \right \rVert} \left \langle \frac{{\bf d}'}{\left \lVert {\bf d}' \right \rVert }, {\bf o}' \right \rangle \right \rVert } \right) =&\\
&= \dv{}{\hat{\alpha}_i} \left( \frac{1}{1 + e^{-\hat{\alpha}_i}} \right) \cdot e^{ -\frac{1}{2} \left \lVert {\bf o}' - \frac{{\bf d}'}{\left \lVert {\bf d}' \right \rVert} \left \langle \frac{{\bf d}'}{\left \lVert {\bf d}' \right \rVert }, {\bf o}' \right \rangle \right \rVert } =&\\
&= - \frac{\dv{}{\hat{\alpha}_i} \left( 1 + e^{-\hat{\alpha}_i} \right) }{ { \left( 1 + e^{-\hat{\alpha}_i} \right) }^2 } \cdot e^{ -\frac{1}{2} \left \lVert {\bf o}' - \frac{{\bf d}'}{\left \lVert {\bf d}' \right \rVert} \left \langle \frac{{\bf d}'}{\left \lVert {\bf d}' \right \rVert }, {\bf o}' \right \rangle \right \rVert } =&\\
&= - \frac{ - e^{-\hat{\alpha}_i} }{ { \left( 1 + e^{-\hat{\alpha}_i} \right) }^2 } \cdot e^{ -\frac{1}{2} \left \lVert {\bf o}' - \frac{{\bf d}'}{\left \lVert {\bf d}' \right \rVert} \left \langle \frac{{\bf d}'}{\left \lVert {\bf d}' \right \rVert }, {\bf o}' \right \rangle \right \rVert } =&\\
&= \frac{ e^{-\hat{\alpha}_i} }{ { \left( 1 + e^{-\hat{\alpha}_i} \right) }^2 } \cdot e^{ -\frac{1}{2} \left \lVert {\bf o}' - \frac{{\bf d}'}{\left \lVert {\bf d}' \right \rVert} \left \langle \frac{{\bf d}'}{\left \lVert {\bf d}' \right \rVert }, {\bf o}' \right \rangle \right \rVert } =&\\
&= \frac{ \left( 1 + e^{-\hat{\alpha}_i} \right) - 1 }{ { \left( 1 + e^{-\hat{\alpha}_i} \right) }^2 } \cdot e^{ -\frac{1}{2} \left \lVert {\bf o}' - \frac{{\bf d}'}{\left \lVert {\bf d}' \right \rVert} \left \langle \frac{{\bf d}'}{\left \lVert {\bf d}' \right \rVert }, {\bf o}' \right \rangle \right \rVert } =&\\
&= \left( \frac{1}{ 1 + e^{-\hat{\alpha}_i} } - \frac{1}{ { \left( 1 + e^{-\hat{\alpha}_i} \right) }^2 } \right) \cdot e^{ -\frac{1}{2} \left \lVert {\bf o}' - \frac{{\bf d}'}{\left \lVert {\bf d}' \right \rVert} \left \langle \frac{{\bf d}'}{\left \lVert {\bf d}' \right \rVert }, {\bf o}' \right \rangle \right \rVert } =&\\
&= \frac{1}{ 1 + e^{-\hat{\alpha}_i} } \left( 1 - \frac{1}{ 1 + e^{-\hat{\alpha}_i} } \right) \cdot e^{ -\frac{1}{2} \left \lVert {\bf o}' - \frac{{\bf d}'}{\left \lVert {\bf d}' \right \rVert} \left \langle \frac{{\bf d}'}{\left \lVert {\bf d}' \right \rVert }, {\bf o}' \right \rangle \right \rVert }
\end{flalign*}
As for the gradient value for the other trainable parameters ${}\Box{} \ne \hat{\alpha}$, we have:
\begin{flalign*}
\dv{\alpha_i}{{{}\Box{}}_i} &=
\dv{}{{{}\Box{}}_i} \left( \frac{1}{1 + e^{-\hat{\alpha}_i}} \cdot e^{ -\frac{1}{2} \left \lVert {\bf o}' - \frac{{\bf d}'}{\left \lVert {\bf d}' \right \rVert} \left \langle \frac{{\bf d}'}{\left \lVert {\bf d}' \right \rVert }, {\bf o}' \right \rangle \right \rVert } \right) =&\\
&= \frac{1}{1 + e^{-\hat{\alpha}_i}} \cdot \dv{}{{{}\Box{}}_i} \left( e^{ -\frac{1}{2} \left \lVert {\bf o}' - \frac{{\bf d}'}{\left \lVert {\bf d}' \right \rVert} \left \langle \frac{{\bf d}'}{\left \lVert {\bf d}' \right \rVert }, {\bf o}' \right \rangle \right \rVert } \right) =&\\
&= - \frac{1}{2} \cdot \frac{1}{1 + e^{-\hat{\alpha}_i}} \cdot e^{ -\frac{1}{2} \left \lVert {\bf o}' - \frac{{\bf d}'}{\left \lVert {\bf d}' \right \rVert} \left \langle \frac{{\bf d}'}{\left \lVert {\bf d}' \right \rVert }, {\bf o}' \right \rangle \right \rVert } \cdot \dv{}{{{}\Box{}}_i} \left( \left \lVert {\bf o}' - \frac{{\bf d}'}{\left \lVert {\bf d}' \right \rVert} \left \langle \frac{{\bf d}'}{\left \lVert {\bf d}' \right \rVert }, {\bf o}' \right \rangle \right \rVert \right)
\end{flalign*}
Since:
\begin{flalign*}
& \left \lVert {\bf o}' - \frac{{\bf d}'}{\left \lVert {\bf d}' \right \rVert} \left \langle \frac{{\bf d}'}{\left \lVert {\bf d}' \right \rVert }, {\bf o}' \right \rangle \right \rVert =&\\
&= \left \langle {\bf o}' - \frac{{\bf d}'}{\left \lVert {\bf d}' \right \rVert} \left \langle \frac{{\bf d}'}{\left \lVert {\bf d}' \right \rVert }, {\bf o}' \right \rangle , {\bf o}' - \frac{{\bf d}'}{\left \lVert {\bf d}' \right \rVert} \left \langle \frac{{\bf d}'}{\left \lVert {\bf d}' \right \rVert }, {\bf o}' \right \rangle \right \rangle =&\\
&= \left \langle {\bf o}', {\bf o}' \right \rangle - 2 \left \langle {\bf o}' , \frac{{\bf d}'}{\left \lVert {\bf d}' \right \rVert} \left \langle \frac{{\bf d}'}{\left \lVert {\bf d}' \right \rVert }, {\bf o}' \right \rangle \right \rangle + \left \langle \frac{{\bf d}'}{\left \lVert {\bf d}' \right \rVert} \left \langle \frac{{\bf d}'}{\left \lVert {\bf d}' \right \rVert }, {\bf o}' \right \rangle , \frac{{\bf d}'}{\left \lVert {\bf d}' \right \rVert} \left \langle \frac{{\bf d}'}{\left \lVert {\bf d}' \right \rVert }, {\bf o}' \right \rangle \right \rangle =&\\
&= \left \langle {\bf o}', {\bf o}' \right \rangle - 2 \left \langle {\bf o}' , \frac{{\bf d}'}{ {\left \lVert {\bf d}' \right \rVert}^2 } \left \langle {\bf d}' , {\bf o}' \right \rangle \right \rangle + \left \langle \frac{{\bf d}'}{ {\left \lVert {\bf d}' \right \rVert}^2 } \left \langle {\bf d}' , {\bf o}' \right \rangle , \frac{{\bf d}'}{ {\left \lVert {\bf d}' \right \rVert}^2 } \left \langle {\bf d}' , {\bf o}' \right \rangle \right \rangle =&\\
&= \left \langle {\bf o}', {\bf o}' \right \rangle - 2 \frac{ { \left \langle {\bf d}' , {\bf o}' \right \rangle }^2 }{\left \langle {\bf d}' , {\bf d}' \right \rangle} + \frac{ { \left \langle {\bf d}' , {\bf o}' \right \rangle }^2 }{\left \langle {\bf d}' , {\bf d}' \right \rangle} =&\\
&= \left \langle {\bf o}', {\bf o}' \right \rangle - \frac{ { \left \langle {\bf d}' , {\bf o}' \right \rangle }^2 }{\left \langle {\bf d}' , {\bf d}' \right \rangle},
\end{flalign*}
we have:
\begin{flalign*}
& \dv{}{{{}\Box{}}_i} \left( \left \lVert {\bf o}' - \frac{{\bf d}'}{\left \lVert {\bf d}' \right \rVert} \left \langle \frac{{\bf d}'}{\left \lVert {\bf d}' \right \rVert }, {\bf o}' \right \rangle \right \rVert \right) =&\\
&= \dv{}{{{}\Box{}}_i} \left( \left \langle {\bf o}', {\bf o}' \right \rangle - \frac{ { \left \langle {\bf d}' , {\bf o}' \right \rangle }^2 }{\left \langle {\bf d}' , {\bf d}' \right \rangle} \right) =&\\
&= \dv{}{{{}\Box{}}_i} \left( \left \langle {\bf o}', {\bf o}' \right \rangle \right) - \dv{}{{{}\Box{}}_i} \left( \frac{ { \left \langle {\bf d}' , {\bf o}' \right \rangle }^2 }{\left \langle {\bf d}' , {\bf d}' \right \rangle} \right) =&\\
&= 2 \left \langle {\bf o}', \dv{{\bf o}'}{{{}\Box{}}_i} \right \rangle - \frac{ \dv{}{{{}\Box{}}_i} \left( { \left \langle {\bf d}', {\bf o}' \right \rangle }^2 \right) \cdot \left \langle {\bf d}', {\bf d}' \right \rangle - { \left \langle {\bf d}', {\bf o}' \right \rangle }^2 \cdot \dv{}{{{}\Box{}}_i} \left( \left \langle {\bf d}', {\bf d}' \right \rangle \right) }{ { \left \langle {\bf d}', {\bf d}' \right \rangle }^2 } =&\\
&= 2 \left \langle {\bf o}', \dv{{\bf o}'}{{{}\Box{}}_i} \right \rangle - \frac{ 2 \cdot \left \langle {\bf d}', {\bf d}' \right \rangle \cdot \left \langle {\bf d}', {\bf o}' \right \rangle \cdot \dv{}{{{}\Box{}}_i} \left( \left \langle {\bf d}', {\bf o}' \right \rangle \right) - 2 \cdot { \left \langle {\bf d}', {\bf o}' \right \rangle }^2 \cdot \left \langle {\bf d}', \dv{{\bf d}'}{{{}\Box{}}_i} \right \rangle }{ { \left \langle {\bf d}', {\bf d}' \right \rangle }^2 } =&\\
&= 2 \left \langle {\bf o}', \dv{{\bf o}'}{{{}\Box{}}_i} \right \rangle - \frac{ 2 \cdot \left \langle {\bf d}', {\bf d}' \right \rangle \cdot \left \langle {\bf d}', {\bf o}' \right \rangle \left( \left \langle {\bf o}', \dv{{\bf d}'}{{{}\Box{}}_i} \right \rangle + \left \langle {\bf d}', \dv{{\bf o}'}{{{}\Box{}}_i} \right \rangle \right) - 2 \cdot { \left \langle {\bf d}', {\bf o}' \right \rangle }^2 \cdot \left \langle {\bf d}', \dv{{\bf d}'}{{{}\Box{}}_i} \right \rangle }{ { \left \langle {\bf d}', {\bf d}' \right \rangle }^2 } =&\\
&= 2 \left \langle {\bf o}', \dv{{\bf o}'}{{{}\Box{}}_i} \right \rangle - 2 \cdot \frac { \left \langle {\bf d}', {\bf o}' \right \rangle }{ \left \langle {\bf d}', {\bf d}' \right \rangle } \cdot \left \langle {\bf d}', \dv{{\bf o}'}{{{}\Box{}}_i} \right \rangle - 2 \cdot \frac { \left \langle {\bf d}', {\bf o}' \right \rangle }{ \left \langle {\bf d}', {\bf d}' \right \rangle } \cdot \left \langle {\bf o}', \dv{{\bf d}'}{{{}\Box{}}_i} \right \rangle + 2 \cdot \frac { \left \langle {\bf d}', {\bf o}' \right \rangle }{ \left \langle {\bf d}', {\bf d}' \right \rangle } \cdot \frac { \left \langle {\bf d}', {\bf o}' \right \rangle }{ \left \langle {\bf d}', {\bf d}' \right \rangle } \cdot \left \langle {\bf d}', \dv{{\bf d}'}{{{}\Box{}}_i} \right \rangle =&\\
&= 2 \left \langle {\bf o}' - {\bf d}' \frac { \left \langle {\bf d}' , {\bf o}' \right \rangle }{ \left \langle {\bf d}', {\bf d}' \right \rangle }, \dv{{\bf o}'}{{{}\Box{}}_i} \right \rangle - 2 \cdot \frac { \left \langle {\bf d}', {\bf o}' \right \rangle }{ \left \langle {\bf d}', {\bf d}' \right \rangle } \cdot \left \langle {\bf o}' - {\bf d}' \frac { \left \langle {\bf d}' , {\bf o}' \right \rangle }{ \left \langle {\bf d}', {\bf d}' \right \rangle }, \dv{{\bf d}'}{{{}\Box{}}_i} \right \rangle =&\\
&= 2 \left \langle {\bf o}' - \frac{{\bf d}'}{ { \left \lVert {\bf d}' \right \rVert }^2 } \left \langle {\bf d}' , {\bf o}' \right \rangle , \dv{{\bf o}'}{{{}\Box{}}_i} \right \rangle - 2 \cdot \frac { \left \langle {\bf d}', {\bf o}' \right \rangle }{ \left \langle {\bf d}', {\bf d}' \right \rangle } \cdot \left \langle {\bf o}' - \frac{{\bf d}'}{ { \left \lVert {\bf d}' \right \rVert }^2 } \left \langle {\bf d}' , {\bf o}' \right \rangle , \dv{{\bf d}'}{{{}\Box{}}_i} \right \rangle =&\\
&= 2 \left \langle {\bf o}' - \frac{{\bf d}'}{ \left \lVert {\bf d}' \right \rVert } \left \langle \frac{{\bf d}'}{ \left \lVert {\bf d}' \right \rVert } , {\bf o}' \right \rangle , \dv{{\bf o}'}{{{}\Box{}}_i} \right \rangle - 2 \cdot \frac { \left \langle {\bf d}', {\bf o}' \right \rangle }{ \left \langle {\bf d}', {\bf d}' \right \rangle } \cdot \left \langle {\bf o}' - \frac{{\bf d}'}{ \left \lVert {\bf d}' \right \rVert } \left \langle \frac{{\bf d}'}{ \left \lVert {\bf d}' \right \rVert } , {\bf o}' \right \rangle , \dv{{\bf d}'}{{{}\Box{}}_i} \right \rangle =&\\
&= 2 \left( \left \langle {\bf o}' - \frac{{\bf d}'}{ \left \lVert {\bf d}' \right \rVert } \left \langle \frac{{\bf d}'}{ \left \lVert {\bf d}' \right \rVert } , {\bf o}' \right \rangle , \dv{{\bf o}'}{{{}\Box{}}_i} \right \rangle - \cdot \frac { \left \langle {\bf d}', {\bf o}' \right \rangle }{ \left \langle {\bf d}', {\bf d}' \right \rangle } \cdot \left \langle {\bf o}' - \frac{{\bf d}'}{ \left \lVert {\bf d}' \right \rVert } \left \langle \frac{{\bf d}'}{ \left \lVert {\bf d}' \right \rVert } , {\bf o}' \right \rangle , \dv{{\bf d}'}{{{}\Box{}}_i} \right \rangle \right)
\end{flalign*}
Now it remains to show how to compute $\dv{{\bf o}'}{{{}\Box{}}_i}$ and $\dv{{\bf d}'}{{{}\Box{}}_i}$. By definition: ${\bf o}' = S_i^{-1} R_i^T \left( {\bf o} - \m_i \right)$ and ${\bf d}' = S_i^{-1} R_i^T {\bf d}$. As for the gradient for the trainable parameter $m_{i,x}$, we have:
$$
\dv{{\bf o}'}{m_{i,x}} = \dv{}{m_{i,x}} \left( S_i^{-1} R_i^T \left( {\bf o} - \m_i \right) \right) = S_i^{-1} R_i^T \dv{}{m_{i,x}} \left( {\bf o} - \m_i \right) = - S_i^{-1} R_i^T \dv{m_i}{m_{i,x}} = - S_i^{-1} R_i^T \varepsilon_x
$$
$$
\dv{{\bf d}'}{m_{i,x}} = \dv{}{m_{i,x}} \left( S_i^{-1} R_i^T {\bf d} \right) = \vec{0},
$$
where $\varepsilon_x$ is the vector of the canonical basis corresponding to the $OX$ axis of the coordinate system. On the other hand for $s_{i,x}$ we obtain:
\begin{flalign*}
\dv{{\bf o}'}{s_{i,x}} &=
\dv{}{s_{i,x}} \left( S_i^{-1} R_i^T \left( {\bf o} - \m_i \right) \right) =&\\
&= \dv{S_i^{-1}}{s_{i,x}} R_i^T \left( {\bf o} - \m_i \right) =&\\
&= \dv{}{s_{i,x}} \left( {
\begin{bmatrix}
\frac{1}{1+e^{-s_{i,x}}} & 0 & 0 \\
0 & \frac{1}{1+e^{-s_{i,y}}} & 0 \\
0 & 0 & \frac{1}{1+e^{-s_{i,z}}}
\end{bmatrix}
}^{-1} \right) R_i^T \left( {\bf o} - \m_i \right) =&\\
&= \dv{}{s_{i,x}} \left( {
\begin{bmatrix}
1+e^{-s_{i,x}} & 0 & 0 \\
0 & 1+e^{-s_{i,y}} & 0 \\
0 & 0 & 1+e^{-s_{i,z}}
\end{bmatrix}
} \right) R_i^T \left( {\bf o} - \m_i \right) =&\\
&= \begin{bmatrix}
\dv{}{s_{i,x}} \left( 1+e^{-s_{i,x}} \right) & 0 & 0 \\
0 & \dv{}{s_{i,x}} \left( 1+e^{-s_{i,y}} \right) & 0 \\
0 & 0 & \dv{}{s_{i,x}} \left( 1+e^{-s_{i,z}} \right)
\end{bmatrix}
R_i^T \left( {\bf o} - \m_i \right) =&\\
&= \begin{bmatrix}
- e^{-s_{i,x}} & 0 & 0 \\
0 & 0 & 0 \\
0 & 0 & 0
\end{bmatrix}
R_i^T \left( {\bf o} - \m_i \right) =&\\
&= - e^{-s_{i,x}} \varepsilon_y \varepsilon_y^T R_i^T \left( {\bf o} - \m_i \right).
\end{flalign*}
Using exactly the same reasoning, we conclude that:
\begin{flalign*}
& \dv{{\bf d}'}{s_{i,x}} = - e^{-s_{i,x}} \varepsilon_y \varepsilon_y^T R_i^T {\bf d} &
\end{flalign*}
The derivation for the $y$ and $z$ components of the above parameters is analogous. Finally, the parameter $q_i = \left( q_{i,r}, q_{i,i}, q_{i,j}, q_{i,k} \right)$ (note the notation conflict: we use $i$ both as the index and as the subscript of the quaternion to indicate that it is the first imaginary component of the quaternion):
$$
\dv{{\bf o}'}{q_{i,{}\Box{}}} = \dv{}{q_{i,{}\Box{}}} \left( S_i^{-1} R_i^T \left( {\bf o} - \m_i \right) \right) = S_i^{-1} \dv{R_i^T}{q_{i,{}\Box{}}} \left( {\bf o} - \m_i \right)
$$
$$
\dv{{\bf d}'}{q_{i,{}\Box{}}} = \dv{}{q_{i,{}\Box{}}} \left( S_i^{-1} R_i^T \bf d \right) = S_i^{-1} \dv{R_i^T}{q_{i,{}\Box{}}} \bf d.
$$
To complete our derivation, we need to provide the formula for $\dv{R_i^T}{q_{i,{}\Box{}}}$. Let's recall that for the quaternion $q_i = \left( q_{i,r}, q_{i,i}, q_{i,j}, q_{i,k} \right)$ which parametrizes the rotation matrix $R_i$, the rotation matrix $R_i$ itself can be obtained from $q_i$ using the following formula:
$$
R = \begin{bmatrix}
1 - cc - dd & bc - ad & bd + ac \\
bc + ad & 1 - bb - dd & cd - ab \\
bd - ac & cd + ab & 1 - bb - cc
\end{bmatrix}
$$
where:
$$
s = \frac{2}{q_{i,r}^2 + q_{i,i}^2 + q_{i,j}^2 + q_{i,k}^2}
$$
$$
\begin{matrix}
bs = q_{i,i} \cdot s & cs = q_{i,j} \cdot s & ds = q_{i,k} \cdot s \\
ab = q_{i,r} \cdot bs & ac = q_{i,r} \cdot cs & ad = q_{i,r} \cdot ds \\
bb = q_{i,i} \cdot bs & bc = q_{i,i} \cdot cs & bd = q_{i,i} \cdot ds \\
cc = q_{i,j} \cdot cs & cd = q_{i,j} \cdot ds & dd = q_{i,k} \cdot ds \\
\end{matrix}
$$
Let us define $aa \coloneq q_{i,r}^2 s$, $a \coloneq q_{i,r}$, $b \coloneq q_{i,i}$, $c \coloneq q_{i,j}$, and $d \coloneq q_{i,k}$. Then, after some computation, we finally obtain:
$$
\dv{R_i^T}{q_{i,r}} = \begin{bmatrix}
s a \left( \left( 1 - aa \right) + \left( 1 - bb \right) \right) &
s \left( -d \left( 1 - aa \right) - ab \cdot c \right) &
s \left( c \left( 1 - aa \right) - ab \cdot d \right) \\
s \left( d \left( 1 - aa \right) - ab \cdot c \right) &
s a \left( \left( 1 - aa \right) + \left( 1 - cc \right) \right) &
s \left( -b \left( 1 - aa \right) - a \cdot cd \right) \\
s \left( -c \left( 1 - aa \right) - ab \cdot d \right) &
s \left( b \left( 1 - aa \right) - a \cdot cd \right) &
s a \left( \left( 1 - aa \right) + \left( 1 - dd \right) \right)
\end{bmatrix}
$$
$$
\dv{R_i^T}{q_{i,i}} = \begin{bmatrix}
s b \left( \left( 1 - bb \right) + \left( 1 - aa \right) \right) &
s \left( c \left( 1 - bb \right) - \left( -ab \cdot d \right) \right) &
s \left( d \left( 1 - bb \right) - ab \cdot c \right) \\
s \left( c \left( 1 - bb \right) - ab \cdot d \right) &
-s b \left( \left( 1 - bb \right) + \left( 1 - dd \right) \right) &
s \left( -a \left( 1 - bb \right) - b \cdot cd \right) \\
s \left( d \left( 1 - bb \right) - \left( -ab \cdot c \right) \right) &
s \left( a \left( 1 - bb \right) - b \cdot cd \right) &
-s b \left( \left( 1 - bb \right) + \left( 1 - cc \right) \right)
\end{bmatrix}
$$
$$
\dv{R_i^T}{q_{i,j}} = \begin{bmatrix}
-s c \left( \left( 1 - cc \right) + \left( 1 - dd \right) \right) &
s \left( b \left( 1 - cc \right) - \left( -a \cdot cd \right) \right) &
s \left( a \left( 1 - cc \right) - b \cdot cd \right) \\
s \left( b \left( 1 - cc \right) - a \cdot cd \right) &
s c \left( \left( 1 - cc \right) + \left( 1 - aa \right) \right) &
s \left( d \left( 1 - cc \right) - \left( -ab \cdot c \right) \right) \\
s \left( -a \left( 1 - cc \right) - b \cdot cd \right) &
s \left( d \left( 1 - cc \right) - ab \cdot c \right) &
-s c \left( \left( 1 - cc \right) + \left( 1 - bb \right) \right)
\end{bmatrix}
$$
$$
\dv{R_i^T}{q_{i,k}} = \begin{bmatrix}
-s d \left( \left( 1 - dd \right) + \left( 1 - cc \right) \right) &
s \left( -a \left( 1 - dd \right) - b \cdot cd \right) &
s \left( b \left( 1 - dd \right) - a \cdot cd \right) \\
s \left( a \left( 1 - dd \right) - b \cdot cd \right) &
-s d \left( \left( 1 - dd \right) + \left( 1 - bb \right) \right) &
s \left( c \left( 1 - dd \right) - \left( -ab \cdot d \right) \right) \\
s \left( b \left( 1 - dd \right) - \left( -a \cdot cd \right) \right) &
s \left( c \left( 1 - dd \right) - ab \cdot d \right) &
s d \left( \left( 1 - dd \right) + \left( 1 - aa \right) \right)
\end{bmatrix}
$$

\subsection{Computing $\dv{\L_{SSIM}}{I_{i,j}}$.}\label{app:grad4}
Recall that:
\begin{flalign*}
\L_{SSIM} &=
\frac{1}{h \cdot w} \sum \limits_{k,l=1}^{h,w} SSIM \left( I_{k,l}, \hat{I}_{k,l} \right) =&\\
&= \frac{1}{h \cdot w} \sum \limits_{k,l=1}^{h,w} \frac{\left( 2 \cdot \mu_{I,k,l} \cdot \mu_{\hat{I},k,l} + c_1 \right) \left( 2 \cdot \sigma_{I, k, l, \hat{I}, k, l} + c_2 \right)}{\left( \mu_{I,k,l}^2 + \mu_{\hat{I},k,l}^2 + c_1 \right) \left( \sigma_{I,k,l}^2 + \sigma_{\hat{I},k,l}^2 + c_2 \right)}
\end{flalign*}
where $c_1 = {\left( k_1{L} \right)}^2 = k_1^2$ and $c_2 = {\left( k_2{L} \right)}^2 = k_2^2$ for $k_1 = 0.01$ and $k_2 = 0.03$, because we assume that the maximum allowed color intensity is $1$. Let us define:
$$ A_{k,l} \coloneq 2 \cdot \mu_{I,k,l} \cdot \mu_{\hat{I},k,l} + c_1 $$
$$ B_{k,l} \coloneq 2 \cdot \sigma_{I, k, l, \hat{I}, k, l} + c_2 $$
$$ C_{k,l} \coloneqq \mu_{I,k,l}^2 + \mu_{\hat{I},k,l}^2 + c_1 $$
$$ D_{k,l} \coloneqq \sigma_{I,k,l}^2 + \sigma_{\hat{I},k,l}^2 + c_2 $$
Then:
\begin{flalign*}
\dv{\L_{SSIM}}{I_{i,j}} &=
\dv{}{I_{i,j}} \left( \frac{1}{h \cdot w} \sum \limits_{k,l=1}^{h,w} \frac{\left( 2 \cdot \mu_{I,k,l} \cdot \mu_{\hat{I},k,l} + c_1 \right) \left( 2 \cdot \sigma_{I, k, l, \hat{I}, k, l} + c_2 \right)}{\left( \mu_{I,k,l}^2 + \mu_{\hat{I},k,l}^2 + c_1 \right) \left( \sigma_{I,k,l}^2 + \sigma_{\hat{I},k,l}^2 + c_2 \right)} \right) =&\\
&= \dv{}{I_{i,j}} \left( \frac{1}{h \cdot w} \sum \limits_{k,l=1}^{h,w} \frac{A_{k,l} \cdot B_{k,l}}{C_{k,l} \cdot D_{k,l}} \right) =&\\
&= \frac{1}{h \cdot w} \sum \limits_{k,l=1}^{h,w} \dv{}{I_{i,j}} \left( \frac{A_{k,l} \cdot B_{k,l}}{C_{k,l} \cdot D_{k,l}} \right) =&\\
&= \frac{1}{h \cdot w} \sum \limits_{k,l=1}^{h,w} \frac{ \dv{}{I_{i,j}} \left( A_{k,l} \cdot B_{k,l} \right) \cdot C_{k,l} \cdot D_{k,l} - A_{k,l} \cdot B_{k,l} \cdot \dv{}{I_{i,j}} \left( C_{k,l} \cdot D_{k,l} \right) }{C_{k,l}^2 \cdot D_{k,l}^2} =&\\
&= \frac{1}{h \cdot w} \sum \limits_{k,l=1}^{h,w} \frac{ \left( \dv{A_{k,l}}{I_{i,j}} \cdot B_{k,l} 
 + A_{k,l} \cdot \dv{B_{k,l}}{I_{i,j}} \right) \cdot C_{k,l} \cdot D_{k,l} - A_{k,l} \cdot B_{k,l} \cdot \left( \dv{C_{k,l}}{I_{i,j}} \cdot D_{k,l} 
 + C_{k,l} \cdot \dv{D_{k,l}}{I_{i,j}} \right) }{C_{k,l}^2 \cdot D_{k,l}^2}
\end{flalign*}
By definition:
\begin{flalign*}
\mu_{I,k,l} &= \sum \limits_{m,n=-r}^r w_{m,n} I_{k+m,l+n} &
\end{flalign*}
\begin{flalign*}
\sigma_{I,k,l}^2 &=
\sum \limits_{m,n=-r}^r w_{m,n} { \left( I_{k+m,l+n} - \mu_{I,k,l} \right) }^2 =&\\
&= \sum \limits_{m,n=-r}^r w_{m,n} \left( I_{k+m,l+n}^2 - 2 \cdot I_{k+m,l+n} \cdot \mu_{I,k,l} + \mu_{I,k,l}^2 \right) =&\\
&= \left( \sum \limits_{m,n=-r}^r w_{m,n} \cdot I_{k+m,l+n}^2\right) - 2 \cdot \mu_{I,k,l} \left( \sum \limits_{m,n=-r}^r w_{m,n} \cdot I_{k+m,l+n} \right) + \mu_{I,k,l}^2 \left( \sum \limits_{m,n=-r}^r w_{m,n} \right) =&\\
&= \left( \sum \limits_{m,n=-r}^r w_{m,n} \cdot I_{k+m,l+n}^2\right) - 2 \cdot \mu_{I,k,l} \cdot \mu_{I,k,l} + \mu_{I,k,l}^2 \cdot 1 =&\\
&= \left( \sum \limits_{m,n=-r}^r w_{m,n} \cdot I_{k+m,l+n}^2\right) - 2 \cdot \mu_{I,k,l}^2 + \mu_{I,k,l}^2 =&\\
&= \left( \sum \limits_{m,n=-r}^r w_{m,n} \cdot I_{k+m,l+n}^2\right) - \mu_{I,k,l}^2 
\end{flalign*}
\begin{flalign*}
&\sigma_{I, k, l, \hat{I}, k, l} =\\
\\
&= \sum \limits_{m,n=-r}^r w_{m,n} \left( I_{k+m,l+n} - \mu_{I,k,l} \right) \left( \hat{I}_{k+m,l+n} - \mu_{\hat{I},k,l} \right) =&\\
&= \sum \limits_{m,n=-r}^r w_{m,n} \left( I_{k+m,l+n} \cdot \hat{I}_{k+m,l+n} - \mu_{\hat{I},k,l} \cdot I_{k+m,l+n} - \mu_{I,k,l} \cdot \hat{I}_{k+m,l+n} + \mu_{I,k,l} \cdot \mu_{\hat{I},k,l} \right) =&\\
\\
&= \left( \sum \limits_{m,n=-r}^r w_{m,n} \cdot I_{k+m,l+n} \cdot \hat{I}_{k+m,l+n} \right) -&\\
&- \mu_{\hat{I},k,l} \left( \sum \limits_{m,n=-r}^r w_{m,n} \cdot I_{k+m,l+n} \right) - \mu_{I,k,l} \left( \sum \limits_{m,n=-r}^r w_{m,n} \cdot \hat{I}_{k+m,l+n} \right) +&\\
&+ \mu_{I,k,l} \cdot \mu_{\hat{I},k,l} \left( \sum \limits_{m,n=-r}^r w_{m,n} \right) =&\\
\\
&= \left( \sum \limits_{m,n=-r}^r w_{m,n} \cdot I_{k+m,l+n} \cdot \hat{I}_{k+m,l+n} \right) - \mu_{\hat{I},k,l} \cdot \mu_{I,k,l} - \mu_{I,k,l} \cdot \mu_{\hat{I},k,l} + \mu_{I,k,l} \cdot \mu_{\hat{I},k,l} \cdot 1 =&\\
&= \left( \sum \limits_{m,n=-r}^r w_{m,n} \cdot I_{k+m,l+n} \cdot \hat{I}_{k+m,l+n} \right) - \mu_{I,k,l} \cdot \mu_{\hat{I},k,l} - \mu_{I,k,l} \cdot \mu_{\hat{I},k,l} + \mu_{I,k,l} \cdot \mu_{\hat{I},k,l} =&\\
&= \left( \sum \limits_{m,n=-r}^r w_{m,n} \cdot I_{k+m,l+n} \cdot \hat{I}_{k+m,l+n} \right) - \mu_{I,k,l} \cdot \mu_{\hat{I},k,l}
\end{flalign*}
where $w_{m, n}$ is the Gaussian window with radius $r$ summing up to $1$ defined for indices: $m, n \in \left \lbrace -r, \ldots, r \right \rbrace \times \left \lbrace -r, \ldots, r \right \rbrace$ and equal to $0$ otherwise. Hence:
\begin{flalign*}
\dv{\mu_{I,k,l}}{I_{i,j}} &=
\dv{}{I_{i,j}} \left( \sum \limits_{m,n=-r}^r w_{m,n} \cdot I_{k+m,l+n} \right) =
\sum \limits_{m,n=-r}^r \dv{}{I_{i,j}} \left( w_{m,n} \cdot I_{k+m,l+n} \right) =
\sum \limits_{m,n=-r}^r w_{m,n} \cdot \delta_{i,k+m} \cdot \delta_{j,l+n} =
w_{i-k,j-l} &
\end{flalign*}
\begin{flalign*}
\dv{\sigma_{I,k,l}^2}{I_{i,j}} &=
\dv{}{I_{i,j}} \left( \left( \sum \limits_{m,n=-r}^r w_{m,n} \cdot I_{k+m,l+n}^2\right) - \mu_{I,k,l}^2 \right) =&\\
&= \left( \sum \limits_{m,n=-r}^r \dv{}{I_{i,j}} \left( w_{m,n} \cdot I_{k+m,l+n}^2 \right) \right) - \dv{\mu_{I,k,l}^2}{I_{i,j}} =&\\
&= \left( \sum \limits_{m,n=-r}^r 2 \cdot w_{m,n} \cdot I_{i,j} \cdot \delta_{i,k+m} \cdot \delta_{j,l+n} \right) - 2 \cdot \mu_{I,k,l} \cdot \dv{\mu_{I,k,l}}{I_{i,j}} =&\\
&= 2 \cdot w_{i-k,j-l} \cdot I_{i,j} - 2 \cdot w_{i-k,j-l} \cdot \mu_{I,k,l}
\end{flalign*}
\begin{flalign*}
\dv{\sigma_{I, k, l, \hat{I}, k, l}}{I_{i,j}} &=
\dv{}{I_{i,j}} \left( \left( \sum \limits_{m,n=-r}^r w_{m,n} \cdot I_{k+m,l+n} \cdot \hat{I}_{k+m,l+n} \right) - \mu_{I,k,l} \cdot \mu_{\hat{I},k,l} \right) =&\\
&= \left( \sum \limits_{m,n=-r}^r \dv{}{I_{i,j}} \left( w_{m,n} \cdot I_{k+m,l+n} \cdot \hat{I}_{k+m,l+n} \right)  \right) - \dv{}{I_{i,j}} \left( \mu_{I,k,l} \cdot \mu_{\hat{I},k,l} \right) =&\\
&= \left( \sum \limits_{m,n=-r}^r w_{m,n} \cdot \hat{I}_{k+m,l+n} \cdot \delta_{i,k+m} \cdot \delta_{j,l+n} \right) - \mu_{\hat{I},k,l} \cdot \dv{ \mu_{I,k,l} }{I_{i,j}} =&\\
&= w_{i-k,j-l} \cdot \hat{I}_{i,j} - w_{i-k,j-l} \cdot \mu_{\hat{I},k,l}
\end{flalign*}
Now we can calculate each of the two summands that appear in the numerator of the fraction in the formula for $\dv{\L_{SSIM}}{I_{i,j}}$:
\begin{flalign*}
& \left( \dv{A_{k,l}}{I_{i,j}} \cdot B_{k,l} + A_{k,l} \cdot \dv{B_{k,l}}{I_{i,j}} \right) \cdot C_{k,l} \cdot D_{k,l} =&\\
&= \left( \dv{}{I_{i,j}} \left( 2 \cdot \mu_{I,k,l} \cdot \mu_{\hat{I},k,l} + c_1 \right) \cdot B_{k,l} + A_{k,l} \cdot \dv{}{I_{i,j}} \left( 2 \cdot \sigma_{I, k, l, \hat{I}, k, l} + c_2 \right) \right) \cdot C_{k,l} \cdot D_{k,l} =&\\
&= \left( 2 \cdot \mu_{\hat{I},k,l} \cdot B_{k,l} \cdot \dv{ \mu_{I,k,l} }{I_{i,j}} + 2 \cdot A_{k,l} \cdot \dv{ \sigma_{I, k, l, \hat{I}, k, l} }{I_{i,j}} \right) \cdot C_{k,l} \cdot D_{k,l} =&\\
&= \left( 2 \cdot \mu_{\hat{I},k,l} \cdot B_{k,l} \cdot w_{i-k,j-l} + 2 \cdot A_{k,l} \cdot \left( w_{i-k,j-l} \cdot \hat{I}_{i,j} - w_{i-k,j-l} \cdot \mu_{\hat{I},k,l} \right) \right) \cdot C_{k,l} \cdot D_{k,l} =&\\
&= \left( 2 \cdot C_{k,l} \cdot D_{k,l} \cdot \mu_{\hat{I},k,l} \left( B_{k,l} - A_{k, l} \right) \right) w_{i-k,j-l} + \hat{I}_{i,j} \left( \left( 2 \cdot A_{k, l} \cdot C_{k,l} \cdot D_{k,l} \right) w_{i-k,j-l} \right)
\end{flalign*}
and
\begin{flalign*}
& A_{k,l} \cdot B_{k,l} \cdot \left( \dv{C_{k,l}}{I_{i,j}} \cdot D_{k,l} + C_{k,l} \cdot \dv{D_{k,l}}{I_{i,j}} \right) =&\\
&= A_{k,l} \cdot B_{k,l} \cdot \left( \dv{}{I_{i,j}} \left( \mu_{I,k,l}^2 + \mu_{\hat{I},k,l}^2 + c_1 \right) \cdot D_{k,l} + C_{k,l} \cdot \dv{}{I_{i,j}} \left( \sigma_{I,k,l}^2 + \sigma_{\hat{I},k,l}^2 + c_2 \right) \right) =&\\
&= A_{k,l} \cdot B_{k,l} \left( 2 \cdot \mu_{I,k,l} \cdot D_{k,l} \cdot \dv{\mu_{I,k,l}}{I_{i,j}} + C_{k,l} \left( 2 \cdot w_{i-k,j-l} \cdot I_{i,j} - 2 \cdot w_{i-k,j-l} \cdot \mu_{I,k,l} \right) \right) =&\\
&= A_{k,l} \cdot B_{k,l} \left( 2 \cdot \mu_{I,k,l} \cdot D_{k,l} \cdot w_{i-k,j-l} + C_{k,l} \left( 2 \cdot w_{i-k,j-l} \cdot I_{i,j} - 2 \cdot w_{i-k,j-l} \cdot \mu_{I,k,l} \right) \right) =&\\
&= \left( 2 \cdot A_{k,l} \cdot B_{k,l} \cdot \mu_{I,k,l} \left( D_{k,l} - C_{k, l} \right) \right) w_{i-k,j-l} + I_{i,j} \left( \left( 2 \cdot A_{k, l} \cdot B_{k,l} \cdot C_{k,l} \right) w_{i-k,j-l} \right)
\end{flalign*}
Let's define:
$$ E_{k, l} \coloneq \frac{2}{C_{k,l}^2 \cdot D_{k,l}^2} \left(  C_{k,l} \cdot D_{k,l} \cdot \mu_{\hat{I},k,l} \left( B_{k,l} - A_{k, l} \right) - A_{k,l} \cdot B_{k,l} \cdot \mu_{I,k,l} \left( D_{k,l} - C_{k, l} \right) \right) $$
$$ F_{k, l} \coloneq \frac{2}{C_{k,l}^2 \cdot D_{k,l}^2} \cdot A_{k, l} \cdot C_{k,l} \cdot D_{k,l} $$
$$ G_{k, l} \coloneq \frac{2}{C_{k,l}^2 \cdot D_{k,l}^2} \cdot A_{k, l} \cdot B_{k,l} \cdot C_{k,l} $$
Substituting the above-mentioned quantities into the formula for $\dv{\L_{SSIM}}{I_{i,j}}$ yields:
\begin{flalign*}
& \dv{\L_{SSIM}}{I_{i,j}} =&\\
&= \frac{1}{h \cdot w} \sum \limits_{k,l=1}^{h,w} \frac{ \left( \dv{A_{k,l}}{I_{i,j}} \cdot B_{k,l} 
 + A_{k,l} \cdot \dv{B_{k,l}}{I_{i,j}} \right) \cdot C_{k,l} \cdot D_{k,l} - A_{k,l} \cdot B_{k,l} \cdot \left( \dv{C_{k,l}}{I_{i,j}} \cdot D_{k,l} 
 + C_{k,l} \cdot \dv{D_{k,l}}{I_{i,j}} \right) }{C_{k,l}^2 \cdot D_{k,l}^2} =&\\
&= \frac{1}{h \cdot w} \sum \limits_{k,l=1}^{h,w} \left( E_{k,l} \cdot w_{i-k,j-l} + \hat{I}_{i,j} \cdot F_{k,l} \cdot w_{i-k,j-l} - I_{i,j} \cdot G_{k,l} \cdot w_{i-k,j-l} \right) =&\\
&= \frac{1}{h \cdot w} \left( \left( \sum \limits_{k,l=1}^{h,w} E_{k,l} \cdot w_{i-k,j-l} \right) + \hat{I}_{i,j} \left( \sum \limits_{k,l=1}^{h,w} F_{k,l} \cdot w_{i-k,j-l} \right) - I_{i,j} \left( \sum \limits_{k,l=1}^{h,w} G_{k,l} \cdot w_{i-k,j-l} \right) \right) =&\\
&= \frac{1}{h \cdot w} \left( { \left( E * w \right) }_{i,j} + \hat{I}_{i,j} { \left( F * w \right) }_{i,j} - I_{i,j} { \left( G * w \right) }_{i,j} \right)
\end{flalign*}
In summary, we have expressed the formula for $\dv{\L_{SSIM}}{I_{i,j}}$ as a weighted sum of three convolutions, which can be computed efficiently using the Fast Fourier Transform.


\end{document}